\documentclass[graybox]{svmult}

\usepackage{type1cm}        
\usepackage{makeidx}         
\usepackage{graphicx}        
\usepackage{multicol}        
\usepackage[bottom]{footmisc}

\usepackage{newtxtext}       %
\usepackage{newtxmath}       

\makeindex             

\usepackage{ulem}
\usepackage[utf8]{inputenc}
\usepackage[export]{adjustbox} 
\usepackage{breakcites}
\usepackage{amsmath}
\usepackage{amsfonts}
\usepackage{amssymb}
\usepackage{bbm}
\usepackage[colorinlistoftodos]{todonotes}
\usepackage[colorlinks=true, allcolors=blue]{hyperref}
\usepackage{natbib}
\usepackage{microtype}
\usepackage{authblk}

\DeclareMathOperator{\median}{median}

\usepackage{caption, subcaption}

\newcommand\blfootnote[1]{%
  \begingroup
  \renewcommand\thefootnote{}\footnote{#1}%
  \addtocounter{footnote}{-1}%
  \endgroup
}

\begin{document}

\title*{
Extracting information from free text through unsupervised graph-based clustering: an application to patient incident records \\
}
\titlerunning{Extracting information from free text}
%
\author{M.\ Tarik Altuncu, Eloise Sorin, Joshua D.\ Symons, Erik Mayer, Sophia N.\ Yaliraki, Francesca Toni, and Mauricio Barahona}
\authorrunning{Altuncu et al.}
\institute{
M.\ Tarik Altuncu \at Department of Mathematics, Imperial College London, 
UK. \\ 
\email{m.altuncu16@imperial.ac.uk}
\and 
Eloise Sorin \at Department of Computing, Imperial College London, 
UK. 
\and 
Joshua D.\ Symons \at Department of Surgery \& Cancer, Imperial College London, 
UK.  
\and 
Erik Mayer \at Department of Surgery \& Cancer, Imperial College London, 
UK. 
\and
Sophia N.\ Yaliraki \at Department of Chemistry, Imperial College London, 
UK. 
\and
Francesca Toni \at Department of Computing, Imperial College London, 
UK. 
\and 
Mauricio Barahona \at Department of Mathematics, Imperial College London, 
UK. 
\\ \email{m.barahona@imperial.ac.uk}
}
\maketitle

\abstract{The large volume of text in electronic healthcare records often remains underused due to a lack of methodologies to extract interpretable content. Here we present an unsupervised framework for the analysis of free text that combines text-embedding with paragraph vectors and graph-theoretical multiscale community detection. 
We analyse text from a corpus of patient incident reports from the National Health Service in England to find content-based clusters of reports in an unsupervised manner and at different levels of resolution. 
Our unsupervised method 
extracts groups with high intrinsic textual consistency and compares well against categories hand-coded by healthcare personnel. 
We also show how to use our content-driven clusters to improve the supervised prediction of the degree of harm of the incident based on the text of the report.
Finally, we discuss future directions to monitor reports over time, and to detect emerging trends outside pre-existing categories. 
}

    \section{Introduction}
    \label{sec:Introduction}

\blfootnote{This chapter contains material from our published manuscript \citep{tarik}
and additional original research. Our published work appeared in an open-access journal, under the terms of the Creative Commons Attribution 4.0 International License (\url{http://creativecommons.org/licenses/ by/4.0/}), which permits unrestricted use, distribution, and reproduction in any medium, provided appropriate credit to the original author(s) and the source is given. 
}

The vast amounts of data collected by healthcare providers in conjunction with modern data analytics present a unique opportunity to improve the quality and safety of medical care for patient benefit~\citep{colijn2017toward}. Much recent research in this area has been on personalised medicine, with the aim to deliver improved diagnostic and treatment through the synergistic integration of datasets at the level of the individual. 
A different source of healthcare data pertains to organisational matters. In the United Kingdom, the National Health Service (NHS) has a long history of documenting the different aspects of healthcare provision, and is currently in the process of making available properly anonymised datasets, with the aim of leveraging advanced analytics to improve NHS services.

One such database is the National Reporting and Learning System (NRLS), a repository of patient safety incident reports from the NHS in England and Wales set up in 2003, which now contains over 13 million records. The incidents are reported under standardised categories and contain both organisational and spatio-temporal information (structured data) and a substantial component of free text (unstructured data) where incidents are described in the `voice' of the person reporting. The incidents are wide ranging: from patient accidents to lost forms or referrals; from delays in admission or discharge to serious untoward incidents, such as retained foreign objects after operations. The review and analysis of such data provides critical insight into complex processes in healthcare with a view towards service improvement. 

Although statistical analyses are routinely performed on the structured data (dates, locations, hand-coded categories, etc),
free text is typically read manually and often ignored in practice, unless a detailed review of a case is undertaken because of the severity of harm that resulted. These limitations are due to  a lack of methodologies that can provide content-based groupings across the large volume of reports submitted nationally for organisational learning.  Automatic categorisation of incidents from free text would sidestep human error and difficulties in assigning incidents to \textit{a priori} pre-defined lists in the reporting system. Such tools can also offer unbiased insight into the root cause analysis of incidents that could improve the safety and quality of care and efficiency of healthcare services.

In this work, we showcase an algorithmic methodology that detects content-based groups of records in an unsupervised manner, based only on the free (unstructured) textual descriptions of the incidents.  
To do so, we combine deep neural-network high-dimensional text-embedding algorithms with graph-theoretical methods for multiscale clustering.  Specifically, we apply the framework of Markov Stability (MS), a multiscale community detection algorithm, to sparsified graphs of documents obtained from text vector similarities. 
Our method departs both from traditional natural language processing tools, which have generally used bag-of-words (BoW) representation of documents and statistical methods based on Latent Dirichlet Allocation (LDA) to cluster documents~\citep{lda}, and from more recent approaches that have used deep neural network based language models, but have used k-means clustering without a graph-based analysis~\citep{Hashimoto2016TopicReviews}. 
Previous applications of network theory to text analysis have included the work of Lanchichinetti and co-workers 
~\citep{PhysRevX.5.011007}, who proposed a probabilistic graph construction analysed with the InfoMap algorithm~\citep{infomap_EPJS}; however, their community detection was carried out at a single-scale and the BoW representation of text lacks the power of text embeddings. 
The application of multiscale community detection allows us to find groups of records with consistent content at different levels of resolution; hence the content categories emerge from the textual data, rather than from pre-designed classifications.
The obtained results can help mitigate human error or effort in finding the right category in complex classification trees. We illustrate in our analysis the insight gained from this unsupervised, multi-resolution approach in this specialised corpus of medical records.

As an additional application, we use machine learning methods for the prediction of the degree of harm of incidents directly from the text in the  NRLS incident reports. Although the degree of harm is recorded by the reporting person for every event, this information can be unreliable as reporters have been known to game the system, or to give different answers depending on their professional status~\citep{riskerror}.
Previous work on predicting the severity of adverse events~\citep{ref2,ref1} used reports submitted to the Advanced Incident Management System
by Australian public hospitals,
and used BoW and Support Vector Machines (SVMs) 
to
detect extreme-risk events. 
Here we demonstrate that 
publicly reported measures
derived from NHS Staff Surveys can help select  
ground truth labels that allow  supervised training
of machine learning classifiers to predict the degree of harm directly from text embeddings. Further, we show that the unsupervised clusters of content derived with our method improve the classification results significantly. 

An \textit{a posteriori} manual labelling by three clinicians agree with our predictions based purely on text almost as much as with the original hand-coded labels. These results indicate that incidents can be automatically classified according to their degree of harm based only on their textual descriptions, and underlines the potential of automatic document analysis to help reduce human workload.

        \subsection*{Data description} 
        \label{sec:Data}

The full dataset includes more than 13 million confidential reports of patient safety incidents reported to the \textit{National Reporting and Learning System} (NRLS) between 2004 and 2016 from NHS trusts and hospitals in England and Wales. Each record has more than 170 features, including organisational details (e.g., time, trust code and location), anonymised patient information, medication and medical devices, among many other details.
In most records, there is also a detailed description of the incident in free text, although the quality of the text is highly variable. 

The records are manually classified by operators according to a two-level system of incident types. The top level contains 15 categories including general classes such as `Patient accident', `Medication', `Clinical assessment', `Documentation', `Admissions/Transfer' or `Infrastructure', alongside more specific groups such as `Aggressive behaviour', `Patient abuse', `Self-harm' or `Infection control'.

Each record is also labelled based on the degree of harm to the patients as one of: `No Harm', `Low Harm', `Moderate Harm', `Severe Harm' or `Death'. 
These degrees are precisely defined by the WHO \citep{who_ICPS} and the NHS \citep{degree_of_harm}.

    \section{Graph-based framework for text analysis and clustering}
    \label{sec:Framework}

Our framework combines text-embedding, geometric graph construction and multi-resolution community detection to identify, rather than impose, content-based clusters from free, unstructured text in an unsupervised manner. 

\begin{figure}[htb]
\centering
\includegraphics[width=\textwidth]{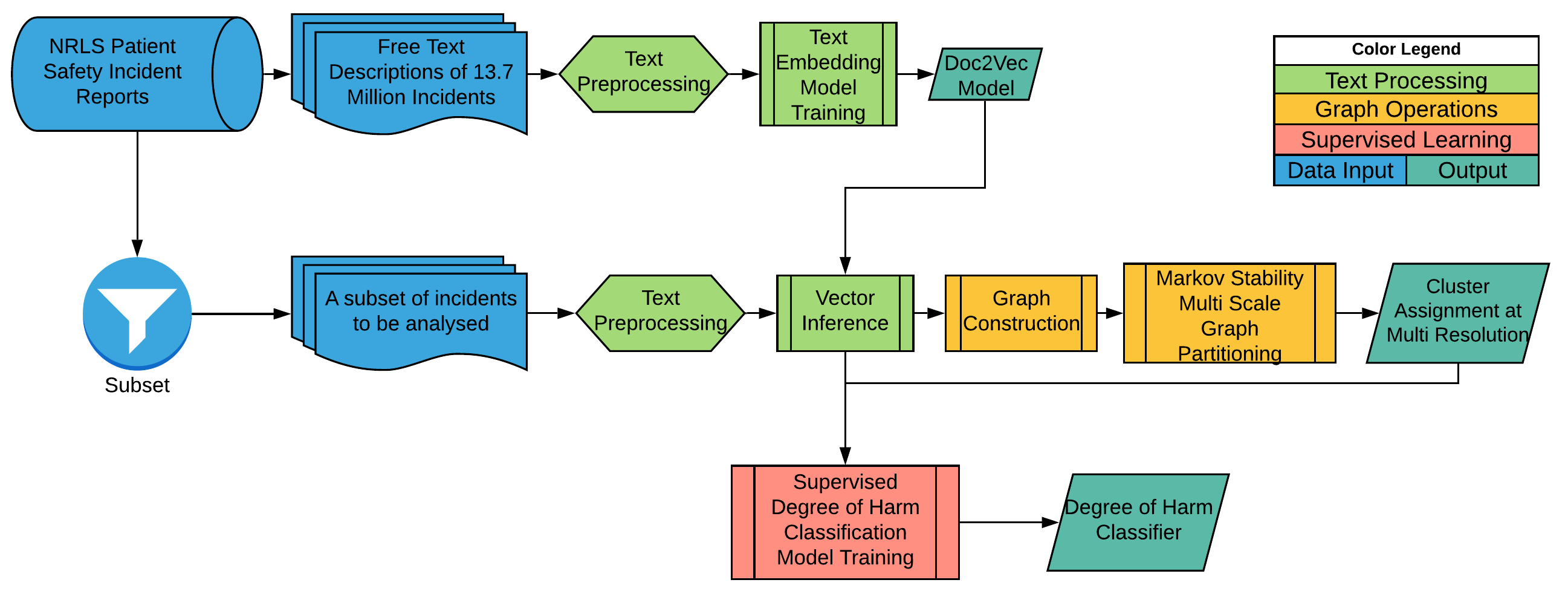}
\caption{
Pipeline for data analysis contains training of the text embedding model along with the two methods we showcase in this work. First is the graph-based unsupervised clustering of documents at different levels of resolution to find topic clusters only from the free text descriptions of hospital incident reports from the NRLS database. Second one uses the topic clusters to improve supervised classification performance of degree of harm prediction.
}
\label{fig:pipeline}
\end{figure}

Figure~\ref{fig:pipeline} shows a summary of our pipeline. First, we pre-process each document to transform text into consecutive word tokens, with words in their most normalised forms and some words removed if they have no distinctive meaning when used out of context~\citep{nltk,porter_old}. 
We then train a paragraph vector model using the Document to Vector (Doc2Vec) framework~\citep{d2v_mikolov} on the full set (13 million) of pre-processed text records. (Training a vector model on smaller sets of 1 million records also produces good results as seen in Table~\ref{table:d2v}). This training step of the text model is only done once.

The trained Doc2Vec model is subsequently used to infer high-dimensional vector descriptions for the text of each document in our target analysis set. 
We then compute a matrix containing all the pairwise (cosine) similarities between the Doc2Vec document vectors. This similarity matrix can be thought of as the adjacency matrix of a full, weighted graph with documents as nodes and edges weighted by their similarity.  We sparsify this graph to the union of a minimum spanning tree and a k-Nearest Neighbors (MST-kNN) graph~\citep{mstknn}, a geometric construction that removes less important similarities but preserves global connectivity for the graph and, hence, for the dataset. 
The MST-kNN graph is then analysed with Markov Stability~\citep{pnasStability,LambiotteMarkovProcess,Delvenne2013,lambiotte_arxiv}, a multi-resolution graph partitioning method that identifies relevant subgraphs (i.e., clusters of documents) at different levels of granularity. MS uses a diffusive process on the graph to reveal the multiscale organisation at different resolutions without the need to choose {\it a priori} the number or type of clusters. 

The partitions found by MS across levels of resolution are analysed {\it a posteriori} through visualisations and quantitative scores. The visualisations include: (i) word clouds to summarise the main content; (ii) graph layouts; and (iii) Sankey diagrams and contingency tables that capture correspondences between partitions. The quantitative scores include: (i) the intrinsic topic coherence (measured by the pairwise mutual information~\citep{pmi_coherence, Newman:2010:AET:1857999.1858011}); and (ii) the similarity to hand-coded categories (measured by the normalised mutual information~~\citep{nmi}).

Our framework also covers prediction of the degree of harm (DoH) caused to the patient usig text embeddings and the unsupervised cluster assignments obtaind from our multiscale graph partitioning.  To perform this task, we use the hand-coded DoH from the NRLS to train three commonly used classifiers~\citep{book, svmeg} (Ridge, Support Vector Machine with a linear kernel, Random Forest) to predict the DoH using TF-iDF and Doc2Vec embeddings of the text  and our MS cluster assignments. The classifiers are then evaluated in predicting the DoH using cross-validation. 

We now explain the steps of the methodological pipeline in more detail.

        \subsection{Text Preprocessing} 
        \label{sec:textPreprocessing}

Text preprocessing is important to enhance the performance of text embedding techniques. We applied standard preprocessing to the raw text of all 13 million records in our corpus, as follows. 
We divide our documents into iterative word tokens using the NLTK library~\citep{nltk} and remove punctuation and digit-only tokens. We then apply word stemming using the Porter algorithm~\citep{porter_old,porter}. If the Porter method cannot find a stemmed version for a token, we apply the Snowball algorithm~\citep{snowball}. Finally, we remove any stop-words (repeat words with low content) using NLTK's stop-word list. Although pre-processing reduces some of the syntactic information, it consolidates the semantic information of the vocabulary. 
We note that the incident descriptions contain typos and acronyms, which have been left uncorrected to avoid manual intervention or the use of spell checkers, so as to mimic as closely as possible a realistic scenario.

        \subsection{Text Vector Embedding} 
        \label{sec:textEmbedding}

Computational text analysis relies on a mathematical representation of the base units of text (character $n$-grams, words or documents). 
Since our methodology is unsupervised, we avoid the use of labelled data, in contrast to supervised or semi-supervised classification methods~\citep{semevalSTS2016, semevalSTS2017}. In our work, we use a representation of text documents as vectors following recent developments in the field. 

Traditionally, bag-of-words (BoW) methods represented documents as vectors of word frequencies weighted by inverse document frequency (TF-iDF).
Such methods provide a statistical description of documents but they do not carry information about the order or proximity of words to each other and hence disregard semantic or syntactic relationships between words. In addition, BoW representations carry little information content as they tend to be high-dimensional and very sparse, due to the large size of word dictionaries and low frequencies of many terms.

Recently, deep neural network language models have successfully overcome the limitations of BoW methods  by incorporating neighbourhoods in the mathematical description of each term.  Distributed Bag of Words (DBOW), better known as Doc2Vec~\citep{d2v_mikolov}, is a form of Paragraph Vectors (PV) which creates a model that represents any word sequence (i.e. sentences, paragraphs, documents) as $d$-dimensional vectors, where $d$ is user-defined (typically $d=300$).
Training a Doc2Vec model starts with a random $d$-dimensional vector assignment for each document in the corpus. A stochastic gradient descent algorithm iterates over the corpus with the objective of predicting a randomly sampled set of words from each document by using only the document's $d$-dimensional vector~\citep{d2v_mikolov}.
The objective function being optimised by PV-DBOW is similar to the skip-gram model in Refs.~\citep{mikolov2013efficient,w2v2}.
Doc2Vec has been shown~\citep{dai2015document} to capture both semantic and syntactic characterisations of the input text, and outperforms BoW-based models such as LDA~\citep{lda}.

\runinhead{Benchmarking the Doc2Vec training:}
Here, we use the Gensim Python library~\citep{gensim} to train the PV-DBOW model.
The Doc2Vec training was repeated several times with a variety of training hyper-parameters (chosen based on our own numerical experiments and the general guidelines provided by~\citep{jhlau}) in order to optimise the output.
To characterise the usability and quality of models, we trained Doc2Vec models using text corpora of different sizes and content with different sets of hyper-parameters. . 
In particular, we checked the effect of corpus size by training Doc2Vec models on the full 13 million NRLS records and on randomly sampled subsets of 1 million and 2 million records. 

Since our target analysis has heavy medical content and specific use of words, we also tested the importance of the training corpus by generating an additional Doc2Vec model using a set of 5 million articles from the English Wikipedia representing standard, generic English usage, which works well in the analysis of news articles~ \citep{altuncu_content-driven_2018}.
 
The results in Table~\ref{table:d2v} show that training on the highly specific text from the NRLS records is an important ingredient in the successful vectorisation of the documents, as shown by the degraded performance for the Wikipedia model across a variety of training hyper-parameters. On the other hand, reducing the size of the corpus from 13 million to 1 million records did not affect the benchmarking dramatically. This robustness of the results to the size of the training corpus was confirmed further with the use of more detailed metrics, as discussed below in Section~\ref{sec:comparisons} (see e.g., Figure~\ref{fig:Corpus_size_SI}).

\begin{table}[htb]
\centering
\caption{Benchmarking of text corpora used for Doc2Vec training. A Doc2Vec model was trained on three corpora of NRLS records of different sizes and a corpus of Wikipedia articles using a variety of hyper-parameters. The scores represent the quality of the vectors inferred using the corresponding model. Specifically, we calcule centroids for the 15 externally hand-coded categories and select the 100 nearest reports for each centroid. We then report the number of incident reports (out of 1500) correctly assigned to their centroid.
}
\label{table:d2v}
\begin{tabular}{|c|c|c||c|c|c|c|}
\hline
\multicolumn{3}{|c||}{\textbf{Hyper-parameters}} & \multicolumn{3}{c|}{\textbf{NRLS}} & \textbf{Wikipedia} \\ \hline
\textbf{\begin{tabular}[c]{@{}c@{}}Window \\ Size\end{tabular}} & \textbf{\begin{tabular}[c]{@{}c@{}}Minimum \\ Count\end{tabular}} & \textbf{Subsampling} & \textbf{1M} & \textbf{2M} & \textbf{13M+} & \textbf{5M+} \\ \hline
15 & 5 & 0.001 & 765 & 755 & \textbf{836} & 531 \\ \hline
5 & 5 & 0.001 & 807 & 775 & 798 & 580 \\ \hline
5 & 20 & 0.001 & 801 & 785 & 809 & 587 \\ \hline
5 & 20 & 0.00001 & - & - & 379 & 465 \\ \hline
15 & 20 & 0.00001 & - & - & 387 & 424 \\ \hline 
\end{tabular}
\end{table}

Based on our benchmarking, henceforth we use the Doc2Vec model trained on the 13+ million NRLS records with the following hyper-parameters: \{training method = dbow, number of dimensions for feature vectors size = 300, number of epochs = 10, window size = 15, minimum count = 5, number of negative samples = 5, random down-sampling threshold for frequent words = 0.001 \}. As an indication of computational cost, the training of this model takes approximately 11 hours (run in parallel with 7 threads) on shared servers.

        \subsection{Similarity graph of documents from text similarities}
        \label{sec:graphConstruction}

Once the Doc2Vec model is trained, we use it to infer a vector for each 
record in our analysis subset and construct $\hat{S}$, a similarity matrix  between the vectors by: computing the matrix of cosine similarities between all pairs of records, $S_\text{cos}$; transforming it into a distance matrix $D_{cos} = 1-S_{cos}$; applying element-wise max norm to obtain $\hat{D}=\|D_{cos}\|_{max}$; and normalising the similarity matrix $\hat{S} = 1-\hat{D}$ which has elements in the interval $[0,1]$.

\begin{figure}[htb!]
\centering
\includegraphics[width=\textwidth]{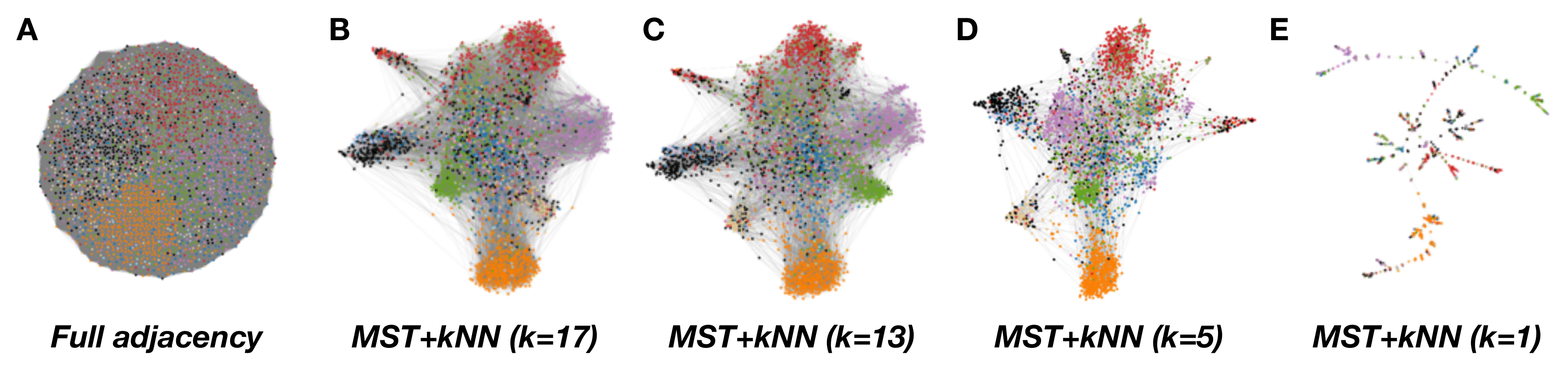}
\caption{
Similarity graphs generated from the vectors of a subset of 3229 patient records. Each node represents a record and is coloured according to its hand-coded, external category to aid visualisation but these external categories are not used to produce our content-driven clustering in Figure~\ref{fig:MS}. 
Layouts for: (a) full, weighted normalised similarity matrix $\hat{S}$ without MST-kNN applied, and \textbf{(b)}--\textbf{(e)}  MST-kNN graphs generated from the data with increasing sparsity as $k$ is reduced.  The structure of the graph is sharpened for intermediate values of $k$. }
\label{fig:fa2_graphConstruction}
\end{figure}

This similarity matrix can be thought of as the adjacency matrix of a fully connected weighted graph. However, such a graph contains many edges with small weights reflecting the fact that in high-dimensional noisy data even the least similar nodes present a substantial degree of similarity. Indeed, such weak similarities are in most cases redundant and can be explained through stronger pairwise similarities.  These weak, redundant edges obscure the graph structure, as shown by the diffuse visualisation in Figure~\ref{fig:fa2_graphConstruction}A.

To reveal the graph structure, we sparsify the similarity matrix to obtain a MST-kNN graph~\citep{mstknn} based on a geometric heuristic that preserves the global connectivity of the graph while retaining details about the local geometry of the dataset. The MST-kNN algorithm starts by computing the minimum spanning tree (MST) of the full matrix $\hat{D}$, i.e., the tree with $(N-1)$ edges connecting all nodes in the graph with minimal sum of edge weights (distances). The MST is computed using the Kruskal algorithm implemented in SciPy~\citep{scipy}. 
To this MST, we add edges connecting each node to its $k$ nearest nodes (kNN) if they are not already in the MST. Here $k$ is an user-defined parameter that regulates the sparsity of the resulting graph. 
The binary adjacency matrix of the MST-kNN graph
is Hadamard-multiplied with $\hat{S}$ to give the adjacency matrix $A$ of the weighted, undirected sparsified graph.

The network visualisations in Figure~\ref{fig:fa2_graphConstruction} give an intuitive picture of the effect of sparsification as $k$ is decreased. 
If $k$ is very small, the graph is very sparse but not robust to noise. As $k$ is increased, the local similarities between documents induce the formation of dense subgraphs (which appear closer in the graph visualisation layout). When the number of neighbours becomes too large, the local structure becomes diffuse and the subgraphs lose coherence, signalling the degradation of the local graph structure. 
Relatively sparse graphs that preserve important edges and global connectivity of the dataset (guaranteed here by the MST) have computational advantages when using community detection algorithms.  

Although we use here the MST-kNN construction due to its simplicity and robustness,
network inference, graph sparsification and graph construction from data is an active area of research, and several alternatives exist based on different heuristics, e.g., Graphical Lasso~\citep{g_lasso}, Planar Maximally Filtered Graph~\citep{pmfg}, spectral sparsification~\citep{spielman2011graph}, or the Relaxed Minimum Spanning Tree (RMST)~\citep{rbs_rmst}. We have experimented with some of those methods and obtained comparable results. A detailed comparison of sparsification methods as well as the choice of distance in defining the similarity matrix $\hat S$ is left for future work.

        \subsection{Multiscale Graph Partitioning}
        \label{sec:MS}

Community detection encompasses various graph partitioning approaches which aim to find `good' partitions into subgraphs (or communities) according to different cost functions, without imposing the number of communities \textit{a priori}~\citep{Schaub2017}. The notion of community depends on the choice of cost function. Commonly, communities are taken to be subgraphs whose nodes are connected strongly within the community with relatively weak inter-community edges. Such structural notion is related to balanced cuts. Other cost functions are posed in terms of transitions inside and outside of the communities, usually as one-step processes~\citep{infomap_EPJS}. When transition paths of all lengths are considered, the concept of community becomes intrinsically multi-scale, i.e., different partitions are relevant at different time scales leading to a multi-level description dictated by the transition dynamics~\citep{pnasStability,Schaub2012ZoomingLens,LambiotteMarkovProcess}. This leads to the framework of Markov Stability (MS), a dynamics-based, multi-scale community detection methodology, which recovers several well-known heuristics as particular cases~\citep{pnasStability,Delvenne2013,lambiotte_arxiv}.

MS is an unsupervised community detection method that finds robust and stable partitions of a graph (and the associated communities) under the evolution of a continuous-time diffusion process without {\it a priori} choice of the number or type of communities or their relative relationships~\citep{pnasStability,Schaub2012ZoomingLens,LambiotteMarkovProcess,ukRiotsTw}~\footnote{The code for Markov Stability is open and accessible at \url{https://github.com/michaelschaub/PartitionStability} and \url{http://wwwf.imperial.ac.uk/~mpbara/Partition_Stability/}, last accessed on March 24, 2018}. In simple terms, MS can be understood by analogy to a drop of ink diffusing on the graph: the ink diffuses homogeneously unless the graph has intrinsic sub-structures, in which case the ink gets transiently contained, over particular time scales, within groups of nodes. The existence of such transients indicates a natural scale to partition the graph along the subgraphs (or communities) where the diffusion is transiently trapped. As the process continues to evolve, the ink diffuses out of those communities but might get transiently contained in other, larger subgraphs, if such multi-level structure exists. By analysing the Markov dynamics over time, MS detects the structure of the graph across scales. If a graph has no natural scales for partitioning, then MS returns no communities.  The Markov time $t$ thus acts as a resolution parameter that allows us to extract robust partitions that persist over particular time scales, in an unsupervised manner.  

Mathematically, given the adjacency matrix $A_{N \times N}$ of the graph obtained as described previously, let us define the diagonal matrix $D=\text{diag}(\mathbf{d})$, 
where $\mathbf{d}=A \mathbf{1}$ is the degree vector.
The random walk Laplacian matrix is defined as $L_\text{RW}=I_N-D^{-1}A$, where $I_N$ is the identity matrix of size $N$ and the transition matrix (or kernel) of the associated continuous-time Markov process is $P(t)=e^{-t L_\text{RW}}, \, t>0$~\citep{LambiotteMarkovProcess}.
Any partition $\mathcal{H}$ into $C$ clusters is associated with a binary membership matrix $H_{N \times C}$ that maps the $N$ nodes into the clusters. Below, we will use the matrix $H$ to denote the corresponding partition $\mathcal{H}$. 
We can then compute the $C\times C$ clustered autocovariance matrix:
\begin{align}
R(t,H) = H^T[\Pi P(t)-\pi\pi^T]H \label{eq:diffusion} 
\end{align}
where $\pi$ is the steady-state distribution of the process and $\Pi=\text{diag}(\pi)$. The element $[R(t,H)]_{\alpha \beta}$ quantifies the probability that a random walker starting from community $\alpha$ at $t=0$ will be in community $\beta$ at time $t$, minus the probability that this event occurs by chance at stationarity.

The above definitions allow us to introduce our cost function measuring the goodness of a partition over time $t$, termed the Markov Stability of partition $H$:
\begin{equation}
r(t,H) = \text{trace} \left[R(t,H)\right]. \label{eq:MS}
\end{equation}
A partition $H$ that maximises $r(t,H)$ is comprised of communities that preserve the flow within themselves over time $t$, since in that case the diagonal elements of $R(t,H)$ will be large and the off-diagonal elements will be small. 
For details, see ~\citep{pnasStability,Schaub2012ZoomingLens,LambiotteMarkovProcess,bacik_celegans}.

Our computational algorithm thus searches for partitions at each Markov time $t$ that maximise $r(t,H)$. Although the maximisation of~\eqref{eq:MS} is an NP-hard problem (hence with no guarantees for global optimality), there are efficient optimisation methods that work well in practice. Our implementation here uses the Louvain Algorithm~\citep{louvain,lambiotte_arxiv} which is efficient and known to give good results when applied to benchmarks. To obtain robust partitions, we run the Louvain algorithm 500 times with different initialisations at each Markov time and pick the best 50 with the highest Markov Stability value $r(t,H)$.  We then compute the variation of information~\citep{Meila2007} of this ensemble of solutions $VI(t)$, as a measure of the reproducibility of the result under the optimisation. In addition, we search for  partitions that are persistent across time $t$, as given by low values of the variation of information between optimised partitions across time $VI(t,t')$. 
Robust partitions are therefore indicated by Markov times where $VI(t)$ shows a dip and $VI(t,t')$ has an extended plateau with low values, indicating consistency under the optimisation and  validity over extended scales~\citep{bacik_celegans,LambiotteMarkovProcess}. 
Below, we apply MS to find partitions across scales of the similarity graph of documents, $A$. The communities detected correspond to groups of documents with similar content at different levels of granularity.

        \subsection{Visualisation and interpretation of the results}
        \label{sec:Visualisation}

\runinhead{Graph layouts:}
We use the ForceAtlas2~\citep{forceAtlas2} layout algorithm to represent graphs on the plane. This layout assigns a harmonic spring to each edge and finds through iterative rearrangements finds an arrangement on the plane that balances attractive and repulsive forces between nodes. Hence similar nodes tend to appear close together on this layout. We colour the nodes by either hand-coded categories (Figure~\ref{fig:fa2_graphConstruction}) or multiscale MS communities (Figure~\ref{fig:MS}). Spatially coherent colourings on this layout imply good clusters in terms of the similarity graph.

\runinhead{Tracking membership through Sankey diagrams:}
Sankey diagrams allow us to visualise the relationship of node membership across different partitions and with respect to the hand-coded categories.
Two-layer Sankey diagrams (e.g., Fig.~\ref{fig:44comms}) reflect the correspondence between MS clusters and the hand-coded external categories, whereas we use a multilayer Sankey diagram in Fig.~\ref{fig:MS} to present the multi-resolution MS community detection across scales.

\runinhead{Normalised contingency tables:}
To capture the relationship between our MS clusters and the hand-coded categories, we also provide a complementary visualisation as z-score heatmaps of normalised contingency tables, e.g., Fig.~\ref{fig:44comms}. This allows us to compare the relative association of content clusters to the external categories at different resolution levels. A quantification of the overall correspondence is also provided by the $NMI$ score in Eq.~\eqref{eq:nmi}.

\runinhead{Word clouds of increased intelligibility through lemmatisation:} 
Our method clusters text documents according to their intrinsic content. This can be understood as a type of topic detection. To visualise the content of clusters, we use Word Clouds as basic, yet intuitive, summaries of information to extract insights and compare \textit{a posteriori} with hand-coded categories. They can also provide an aid for monitoring results when used by practitioners. 

The stemming methods described in Section~\ref{sec:textPreprocessing} truncate words severely to enhance the power of the language processing computational methods by reducing the redundancy in the word corpus. Yet when presenting the results back to a human observer, it is desirable to report the cluster content with words that are readily comprehensible.  To generate comprehensible word clouds in our \textit{a posteriori} analyses, we use a text processing method similar to the one described in~\citep{wordClouds}. Specifically, we use the part of speech (POS) tagging module from NLTK to leave out sentence parts except the adjectives, nouns, and verbs. We also remove less meaningful common verbs such as `be', `have', and `do' and their variations.
The remaining words are then lemmatised in order to normalise variations of the same word. Finally, we use the Python library wordcloud\footnote{The word cloud generator library for Python is open and accessible at \url{https://github.com/amueller/word\_cloud}, last accessed on March 25, 2018} to create word clouds with 2 or 3-gram frequency list of common word groups.

        \subsection{Quantitative benchmarking of topic clusters}
        \label{sec:Benchmarking}

Although our dataset has a classification hand-coded by a human operator, we do not use it in our analysis. Indeed, one of our aims is to explore the relevance of the fixed external classes as compared to content-driven groupings obtained in an unsupervised manner. Therefore we provide a double route to quantify the quality of the clusters by computing two complementary measures:  (i) an intrinsic measure of topic coherence, and (ii) a measure of similarity to the external hand-coded categories.

\runinhead{Topic coherence of text:} 
As an \textit{intrinsic} measure of consistency of word association, we use the \textit{pointwise mutual information} ($PMI$)~\citep{pmi_coherence, pmi_coherence2}. The $PMI$ is an information-theoretical score that captures the probability of words being used together in the same group of documents. The $PMI$ score for a pair of words $(w_1,w_2)$ is:
\begin{equation}
PMI(w_1,w_2)=\log{\frac{P(w_1 w_2)}{P(w_1)P(w_2)}}
\label{eq:pmi}
\end{equation}
where the probabilities of the words $P(w_1)$, $P(w_2)$, and of their co-occurrence $P(w_1 w_2)$ are obtained from the corpus.
We obtain an aggregate $\widehat{PMI}$ for the graph partition $C=\{c_i\}$ by computing the $PMI$ for each cluster, as the median $PMI$ between its 10 most common words (changing the number of words gives similar results), and computing the weighted average of the $PMI$ cluster scores:
\begin{align}
\widehat{PMI} (C) = \sum_{c_i \in C} \frac{n_i}{N} \, \mathop{\median}_{\substack{w_k, w_\ell \in S_i \\ k<\ell}}  PMI(w_k,w_\ell),
\label{eq:pmi_partition}
\end{align}
where $c_i$ denotes the clusters in partition $C$, each with size $n_i$, so that $N=\sum_{c_i \in C} n_i$ is the total number of nodes. Here $S_i$ denotes the set of top 10 words for cluster $c_i$.

The $PMI$ score has been shown to perform well~\citep{pmi_coherence, pmi_coherence2} when compared to human interpretation of topics on different corpora~\citep{pmi_coherence_lda, twitter_pmi_coherence}, and is designed to evaluate topical coherence for groups of documents, in contrast to other tools aimed at short forms of text. See~\citep{semevalSTS2016, semevalSTS2017, semeval2016Samsung, semaval2017ECNU} for other examples. 

Here, we use the $\widehat{PMI}$ score to evaluate partitions without 
any reference to an externally labelled `ground truth'. 

\runinhead{Similarity between the obtained partitions and the hand-coded categories:} 
To quantify how our content-driven unsupervised clusters compare against the external classification, we use the normalised mutual information ($NMI$), a well-known information-theoretical score that quantifies the similarity between clusterings considering correct and incorrect assignments in terms of the information between the clusterings.  The NMI between two partitions $C$ and $D$ of the same graph is:
\begin{equation}
\label{eq:nmi}
NMI(C,D)=\frac{I(C,D)}{\sqrt{H(C)H(D)}}=\frac{\sum\limits_{c \in C} \sum\limits_{d \in D} p(c,d) \, \log\dfrac{p(c,d)}{p(c)p(d)}}{\sqrt{H(C)H(D)}}
\end{equation}
where $I(C,D)$ is the Mutual Information and $H(C)$ and $H(D)$ are the entropies of the two partitions. 

The $NMI$ is bounded ($0 \leq NMI \leq 1$) and a higher value corresponds to higher similarity of the partitions (i.e., $NMI=1$ when there is perfect agreement between partitions $C$ and $D$). The $NMI$ score is directly related\footnote{\url{http://scikit-learn.org/stable/modules/generated/sklearn.metrics.v\_measure\_score.html}} to the \textit{V-measure} in the computer science literature~\citep{vmeasure}.

       \subsection{Supervised Classification for Degree of Harm}
        \label{sec:featureExtraction}

As a further application of our work, we have carried out a supervised classification task aimed at predicting the degree of harm of an incident directly from the text and the hand-coded features (e.g., external category, medical specialty, location). 
A one-hot encoding is applied to turn these categorical values into numerical ones. 
We also checked if using our unsupervised content-driven cluster labels as additional features can improve the performance of the supervised classification.

The supervised classification was carried out by training on features and text three classifiers commonly applied to text classification tasks~\citep{book,svmeg}: a Ridge classifier, Support Vector Machines with a linear kernel, and Random Forests. 
The goal is to predict the degree of harm (DoH) among five possible values (1-5). 
The classification is carried out with five-fold cross validation, using 80\% of the data to train the model and the remaining 20\% to test it. As a measure of performance of the classifiers and models, we use the weighted average of the F1 score for all levels of DoH, which takes into account both precision and recall, i.e., both the exactness and completeness of the model.

    \section{Application to the clustering of hospital incident text reports} 
    \label{sec:Application}
    
We showcase our methodology through the analysis of the text from NRLS patient incident reports. In addition to textual descriptions, the reports are hand-coded upon reporting with up to 170 features per case, including a two-level manual classification of the incidents. 

Here, we only use the text component and apply our graph-based text clustering to a set of 3229 reports from St Mary's Hospital, London (Imperial College Healthcare NHS Trust) over three months in 2014. 
As summarised in Figure~\ref{fig:pipeline}, we start by training our Doc2Vec text embedding using the full 13+ million records collected by the NRLS since 2004 
(although, as discussed above, a much smaller corpus of NRLS documents can be used).
We then infer vectors for our 3229 records, compute the cosine similarity matrix and construct an MST-kNN graph with $k=13$ for our graph-based clustering. 
(We have confirmed the robustness of the MST-kNN construction in our data for $k>13$ by scanning values of $k \in [1,50]$, see Section~\ref{sec:comparisons}). 
We then applied Markov Stability, a multi-resolution graph partitioning algorithm to the MST-kNN graph. We scan across Markov time ($t \in [0.01, 100]$ in steps of 0.01). At each $t$, we run 500 independent Louvain optimisations to select the optimal partition found, as well as 
quantifying the robustness to optimisation by computing the average variation of information $VI(t)$ between the top 50 partitions. 
Once the full scan across $t$ is finalised,
we compute $VI(t,t')$, the variation of information between the optimised partitions found across the scan in Markov time, to select partitions that are robust across scales.

        \subsection{Markov Stability extracts content clusters at different levels of granularity}
        \label{sec:MSAnalysis}

\begin{figure}[htb!]
\centering
\includegraphics[height=.4\paperheight]{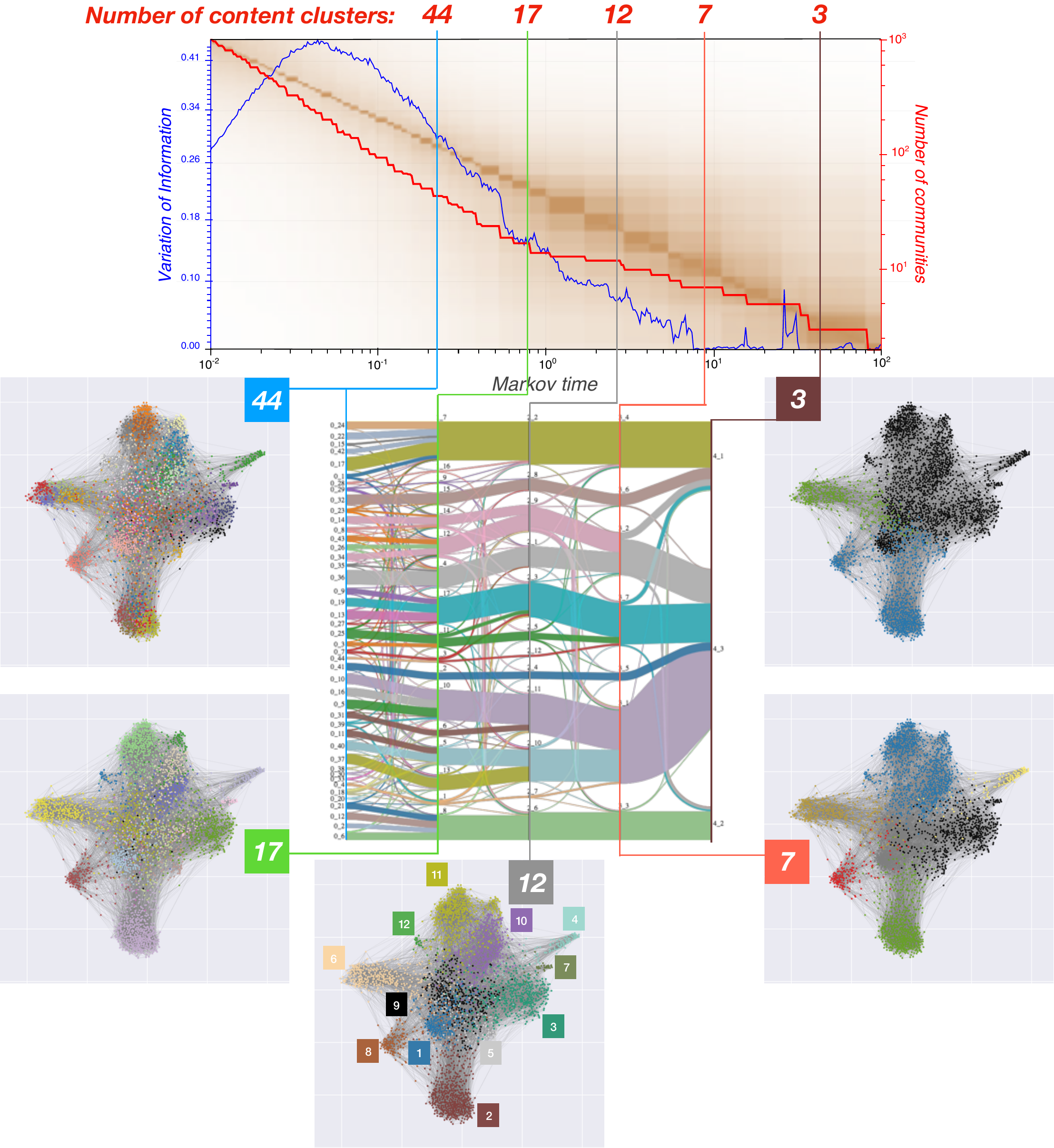}
\caption{
The top plot presents the results of the Markov Stability algorithm across Markov times, showing the number of clusters of the optimised partition (red), the variation of information $VI(t)$ for the ensemble of optimised solutions at each time (blue) and the variation of Information $VI(t,t')$ between the optimised partitions across Markov time (background colourmap). Relevant partitions are indicated by dips of $VI(t)$ and extended plateaux of $VI(t,t')$. We choose five levels with different resolutions (from 44 communities to 3) in our analysis. The Sankey diagram below illustrates how the communities of documents (indicated by numbers and colours) map across Markov time scales. The community structure across scales present a strong quasi-hierarchical character---a result of the analysis and the properties of the data, since it is not imposed \textit{a priori}. The different partitions for the five chosen levels are shown on a graph layout for the document similarity graph created with the MST-kNN algorithm with $k=13$. The colours correspond to the communities found by MS indicating content clusters.}
\label{fig:MS}
\end{figure}
 
Figure~\ref{fig:MS} presents a summary of our MS analysis. We plot the number of clusters of the optimal partition and the two metrics of variation of information across all Markov times. The existence of a long plateau in $VI(t,t')$ coupled to a dip in $VI(t)$ implies the presence of a partition that is robust both to the optimisation and across Markov time. 
To illustrate the multi-scale features of the method, we choose several of these robust partitions, from finer (44 communities) to coarser (3 communities), obtained at five Markov times and examine their structure and content. The multi-level Sankey diagram summarises the relationship of the partitions across levels. 

The MS analysis of the graph reveals a multi-level structure of partitions, with a strong quasi-hierarchical organisation. We remark that our optimisation does not impose any hierarchical structure \emph{a priori}, so that the observed consistency of communities across levels is intrinsic to the data and suggests the existence of sub-themes that integrate into larger thematic categories. The unsupervised detection of intrinsic scales by MS enables us to obtain groups of records with high content similarity at different levels of granularity. This capability can be used by practitioners to tune the level of description to their specific needs, and is used below as an aid in our supervised classification task in Section~\ref{sec:ApplicationEloise}.

\begin{figure}[htb!]
\centering
\includegraphics[width=\textwidth]{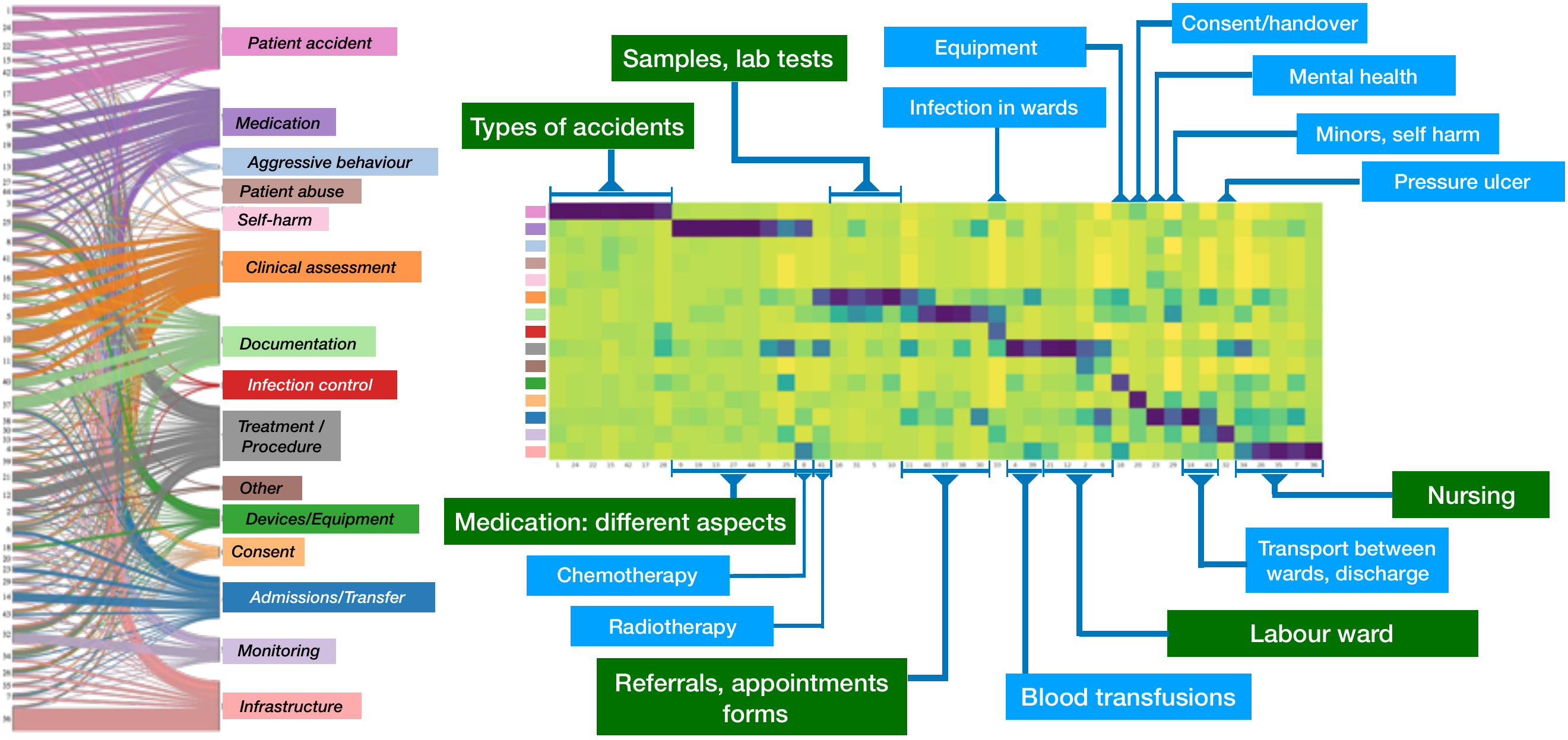}
\caption{Summary of the 44-community partition found with the MS algorithm in an unsupervised manner directly from the text of the incident reports. The 44 content communities are compared \textit{a posteriori} to the 15 hand-coded categories (indicated by names and colours) through a Sankey diagram between communities and categories (left), and through a z-score contingency table (right). We have assigned a descriptive label to the content communities based on their word clouds in Figure~\ref{fig:44comms_SI}.
}
\label{fig:44comms}
\end{figure}

\begin{figure}[htb!]
\centering
\includegraphics[width=\textwidth]{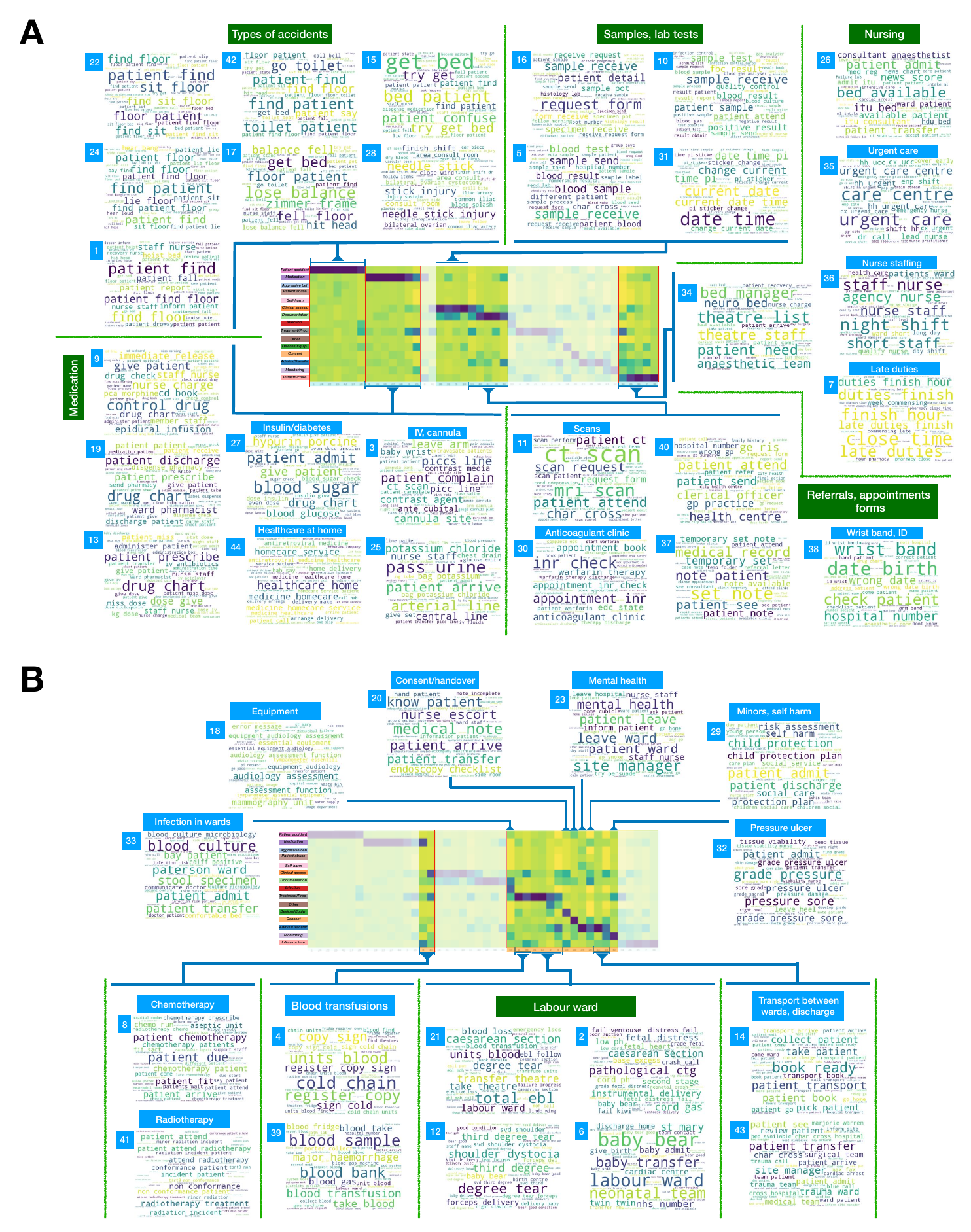}
\caption[Word clouds for the 44 community partitions]{
Word clouds of the 44-community partition showing the detailed content of the communities found. The word clouds are split into two sub-figures (A) and (B) for ease of visualisation.
}
\label{fig:44comms_SI}
\end{figure}

\begin{figure}[htbp!]
\centering
\includegraphics[width=\textwidth]{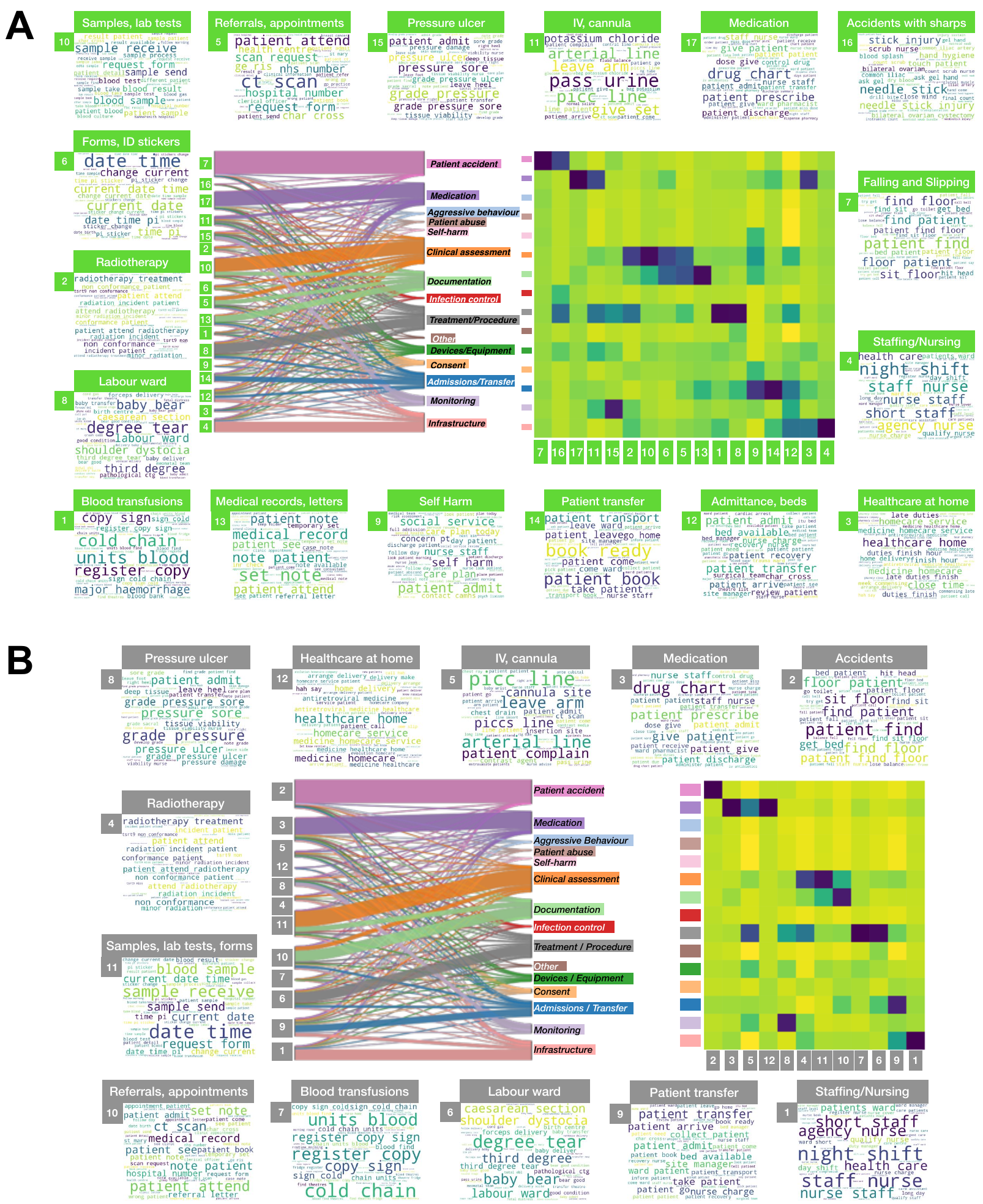}

\caption{Summary of: (A) 17-way and (B) 12-way MS content clusters and their correspondence to the external categories. 
}
\label{fig:12-17comms}
\end{figure}

\begin{figure}[htbp!]
\centering
\includegraphics[width=1\textwidth]{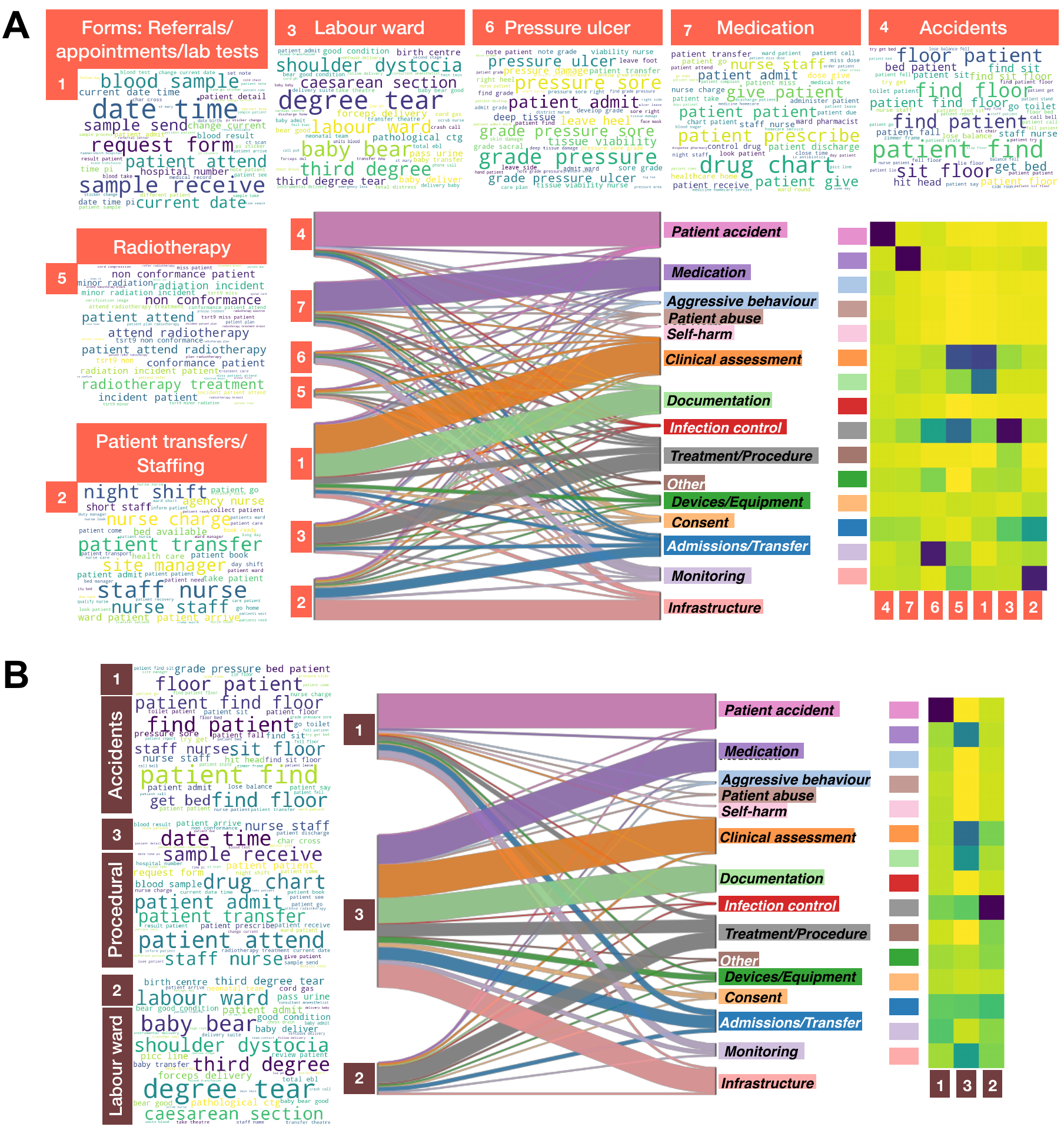}
\caption[Results of coarser MS partitions]{
Summary of MS partitions into (A) 7 communities and (B) 3 communities, showing their correspondence to external hand-coded categories. Some of the MS content clusters have strong medical content (e.g., \textit{labour ward, radiotherapy, pressure ulcer}) and are not grouped with other procedural records due to their semantic distinctiveness, even to this coarse level of clustering.   
}
\label{fig:7_3comms}
\end{figure}

To ascertain the relevance of the layers of content found by MS, we examined the five levels of resolution in Figure~\ref{fig:MS}. For each level, we produced lemmatised word clouds, which we used to generate descriptive content labels for the communities. We then compared \textit{a posteriori}
the content clusters with the hand-coded categories through a Sankey diagram and a contingency table.
The results are shown in Figures~\ref{fig:44comms}--\ref{fig:7_3comms} 
for each of the levels.

The partition into 44 communities presents content clusters with well-defined characterisations, as shown by the Sankey diagram and the highly clustered structure of the contingency table (Figure~\ref{fig:44comms}). 
Compared to the 15 hand-coded categories, this 44-community partition provides finer groupings corresponding to specific sub-themes within the generic hand-coded categories. This is apparent in the hand-coded classes `Accidents', `Medication', `Clinical assessment', `Documentation' and `Infrastructure', where a variety of meaningful subtopics are identified (see Fig.~\ref{fig:44comms_SI} for details).
In other cases, however, the content clusters cut across the external categories, e.g., the clusters on \textit{labour ward, chemotherapy, radiotherapy and infection control} are coherent in content but can belong to several of the external classes. 
At this level of resolution, our algorithm also identified highly specific topics as separate content clusters, including \textit{blood transfusions, pressure ulcer, consent, mental health}, and \textit{child protection}, which have no direct relationship with the external classes provided to the operator.

Figure~\ref{fig:12-17comms}A and \ref{fig:12-17comms}B present the results for two partitions at medium level of resolution, where the number of communities (12 and 17) is close to that of hand-coded categories (15). 
As expected from the quasi-hierarchy detected by our multi-resolution analysis, we find that the communities in the 17-way and 12-way partitions emerge from consistent aggregation of the smaller communities in the 44-way partition in Figure~\ref{fig:44comms}. Focussing on the 12-way partition, we see that some of the sub-themes in Figure~\ref{fig:44comms_SI} are merged into more general topics. An example is \textit{Accidents} (community 2 in Fig.~\ref{fig:12-17comms}A), a merger of seven finer communities, which corresponds well with the external category `Patient accidents'. A similar phenomenon is seen for the \textit{Nursing} cluster (community 1), which falls completely under the external category `Infrastructure'. The clusters related to `Medication' similarly aggregate into a larger community (community 3), yet there still remains a smaller, specific community related to \textit{Homecare medication} (community 12) with distinct content. 
Other communities, on the other hand, still strand across external categories. This is clearly observable in communities 10 and 11 (\textit{Samples/ lab tests/forms} and \textit{Referrals/appointments}), which fall naturally across the `Documentation' and `Clinical Assessment'. Similarly, community 9 (\textit{Patient transfers}) sits across the `Admission/Transfer' and `Infrastructure' external categories, due to its relation to nursing and hospital constraints. 
A substantial proportion of records was hand-coded under the generic `Treatment/Procedure' class, yet MS splits into into content clusters that retain medical coherence, e.g., \textit{Radiotherapy} (Comm.~4), \textit{Blood transfusions} (Comm.~7), \textit{IV/cannula} (Comm.~5), \textit{Pressure ulcer} (Comm.~8), and the large community \textit{Labour ward} (Comm.~6).

The medical specificity of the \textit{Radiotherapy}, \textit{Pressure ulcer} and \textit{Labour ward} clusters means that they are still preserved as separate groups to the next level of coarseness in the 7-way partition (Figure~\ref{fig:7_3comms}A). The mergers in this case lead to a larger communities referring to \textit{Medication}, \textit{Referrals/Forms} and \textit{Staffing/Patient~transfers}. Figure~\ref{fig:7_3comms}B shows the final level of agglomeration into 3 content clusters: records referring to \textit{Accidents}; a group broadly referring to matters \textit{Procedural} (referrals, forms, staffing, medical procedures) cutting across external categories; and the \textit{Labour ward} cluster, still on its own as a subgroup with distinctive content.

\begin{figure}[bp]
\centering
\includegraphics[height=\textwidth,angle=90]{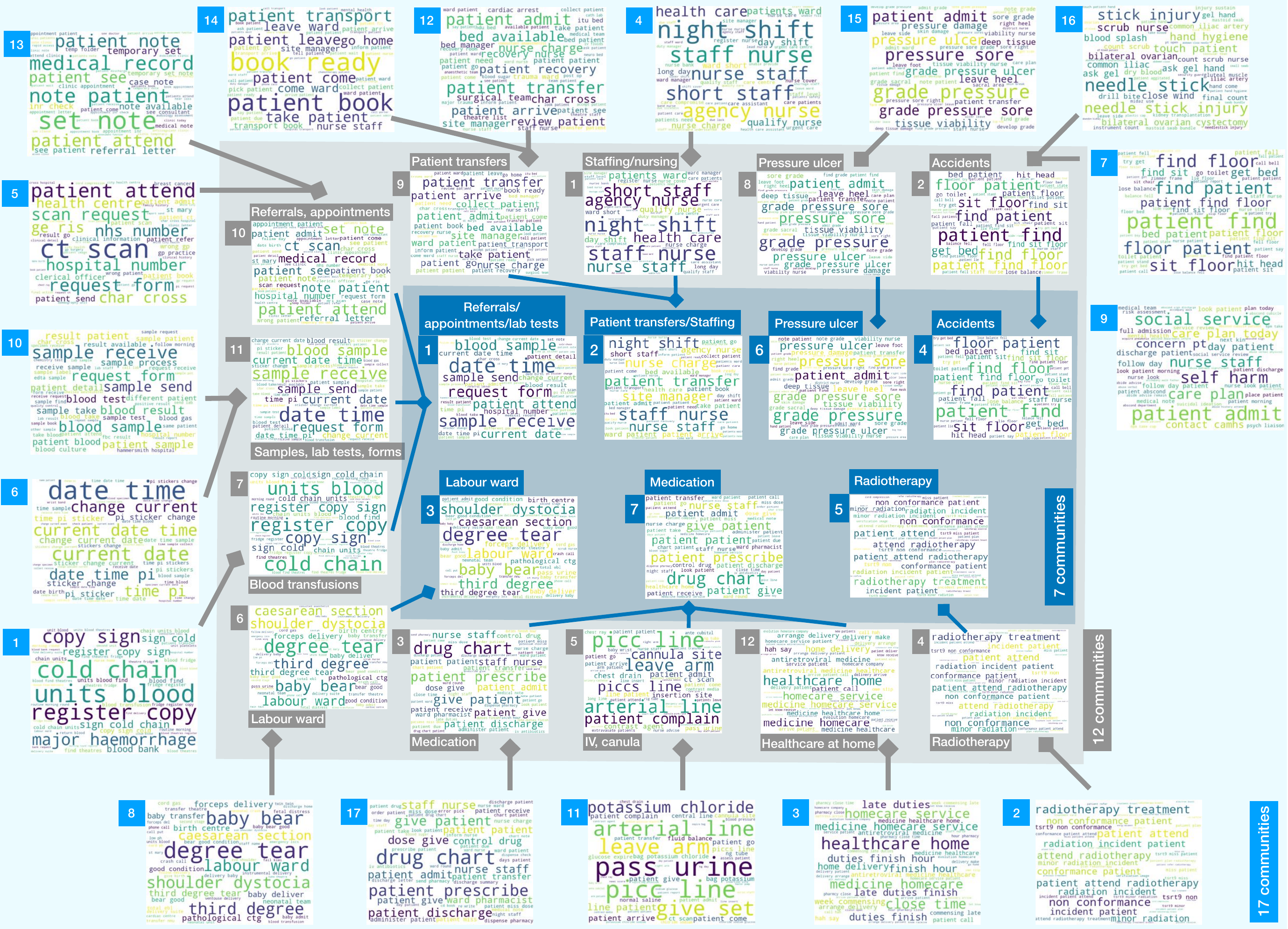}
\caption{
Word clouds of the MS partitions into 17, 12 and 7 clusters show a multi-resolution coarsening in the content following the quasi-hierarchical community structure found in the document similarity graph.
} 
\label{fig:17_12_7_text}
\end{figure}

This process of agglomeration of content, from sub-themes into larger themes, as a result of the multi-scale hierarchy of MS graph partitions is shown explicitly with word clouds in Figure~\ref{fig:17_12_7_text} for the 17-, 12- and 7-way partitions.
Our results show good overall correspondence with the hand-coded categories across resolutions, yet our results also reveal complementary categories of incidents not defined in the external classification. The possibility of tuning the granularity afforded by our method can be used to provide a distinct level of resolution in certain areas corresponding to specialised or particular sub-themes.

        \subsection{Robustness of the results and comparison with other methods}
\label{sec:comparisons}

We have examined quantitatively the robustness of the results to parametric and methodological choices in different steps of our framework. Specifically, we evaluate the effect of: (i) using Doc2Vec embeddings instead of BoW vectors; (ii) the size of corpus for training Doc2Vec; (iii) the sparsity of the MST-kNN graph construction.  We have also carried out quantitative comparisons to other methods for topic detection and clustering: (i) LDA-BoW, and (ii) several standard clustering methods.

\begin{figure}[htb]
\centering
\includegraphics[width=0.9\textwidth,angle=0]{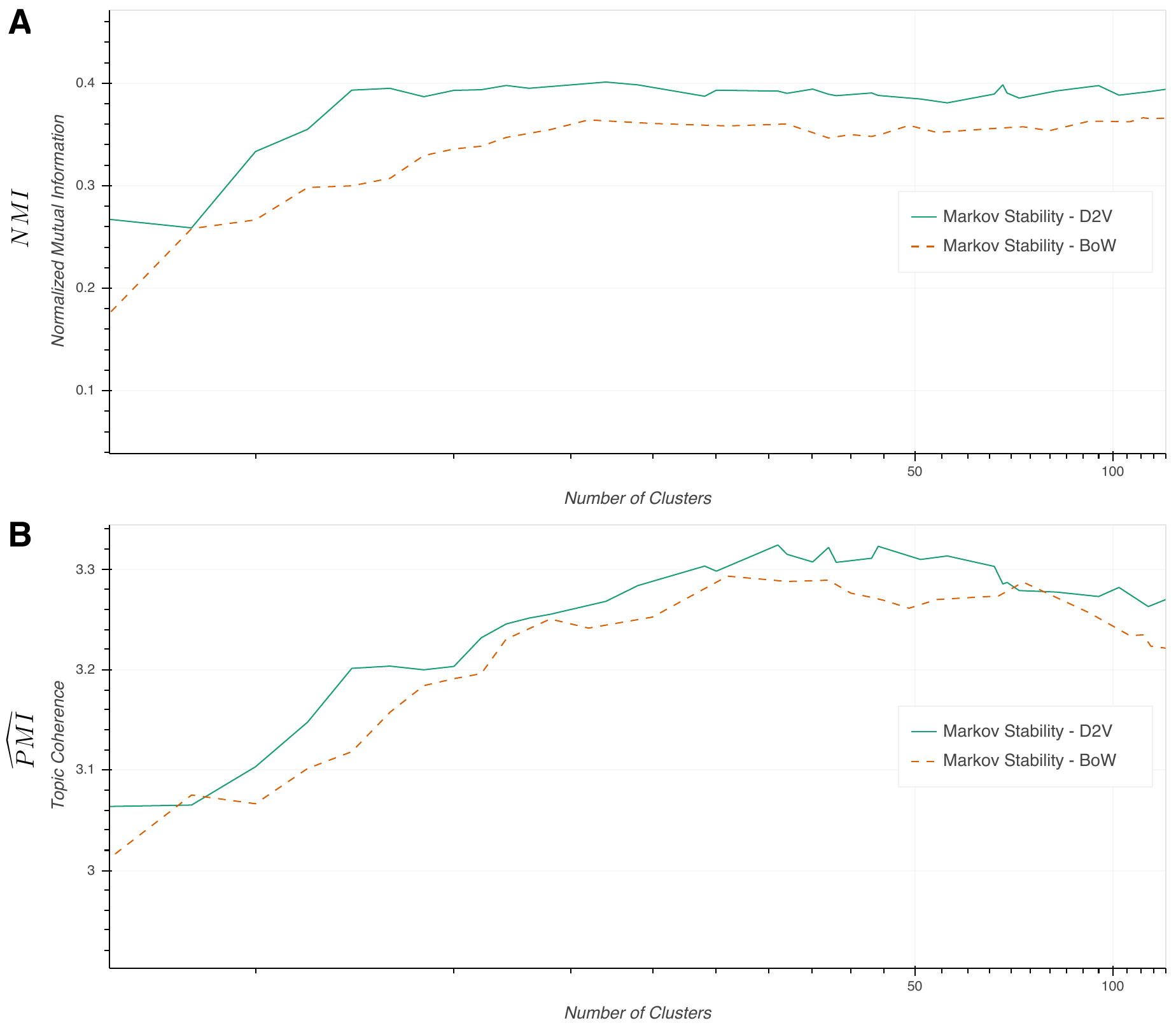}
\caption{
Comparison of MS applied to Doc2Vec versus BoW (using TF-iDF) similarity graphs obtained under the same graph construction.
(A) Similarity against the externally hand-coded categories measured with $NMI$; (B) intrinsic topic coherence of the computed clusters measured with $\widehat{PMI}$.}
\label{fig:D2VvsBoW_MS}
\end{figure}

\runinhead{Doc2Vec provides improved clusters compared to BoW:}
As compared to standard bag of words (BoW), fixed-sized vector embeddings (Doc2Vec) produces lower dimensional vector representations with higher semantic and syntactic content. Doc2Vec outperforms BoW representations in practical benchmarks of semantic similarity and is less sensitive to hyper-parameters~\citep{dai2015document}.
To quantify the improvement provided by Doc2Vec, we constructed a MST-kNN graph from TF-iDF vectors and ran MS on this TF-iDF similarity graph. Figure~\ref{fig:D2VvsBoW_MS} shows that Doc2Vec outperforms BoW across all resolutions in terms of both $NMI$ and $\widehat{PMI}$ scores.

\begin{figure}[htb]
\centering
\includegraphics[width=0.9 \textwidth]{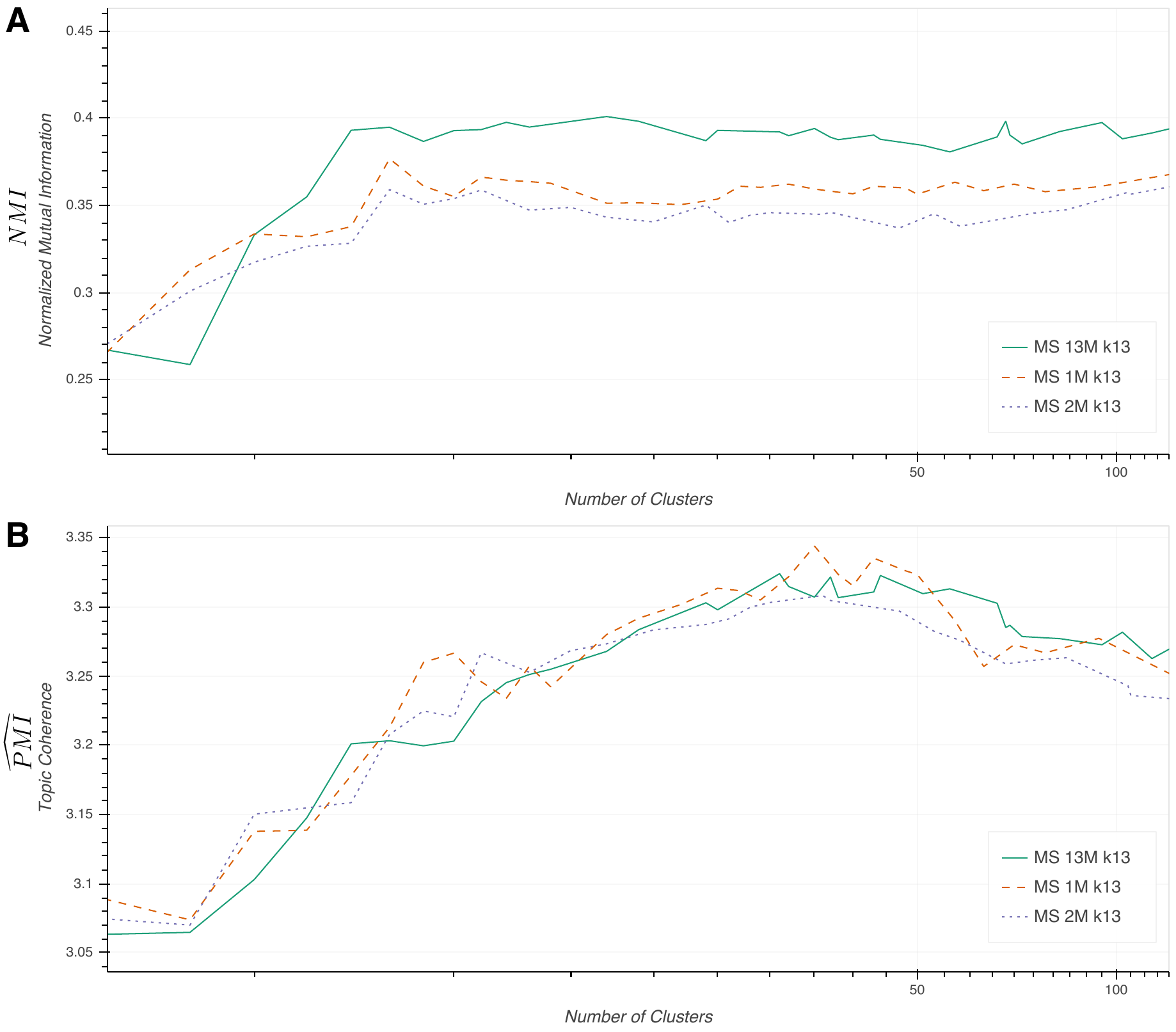}
 \caption{Evaluating the effect of the size of the training corpus. (A) Similarity to hand-coded categories (measured with $NMI$) and (B) Topic Coherence score (measured with $\widehat{PMI}$) of the MS clusters when the Doc2Vec model is trained on: 1 million, 2 million, and the full set of 13 million records. The corpus size does not affect the results.
 }
\label{fig:Corpus_size_SI}
\end{figure}

\runinhead{Robustness to the size of the Doc2Vec training dataset :}
Table~\ref{table:d2v} indicates a small effect of the size of the training corpus on the Doc2Vec model. To confirm this, we trained two additional Doc2Vec models on sets of 1 million and 2 million records (randomly chosen from the full 13+ million records) and followed the same procedure to construct the MST-kNN graph and carry out the MS analysis. 
Figure~\ref{fig:Corpus_size_SI}
shows that the performance is affected only mildly by the size of the Doc2Vec training set.

\runinhead{Robustness to the level of graph sparsification:}

\begin{figure}[htbp]
\centering
\includegraphics[width=0.9\textwidth]{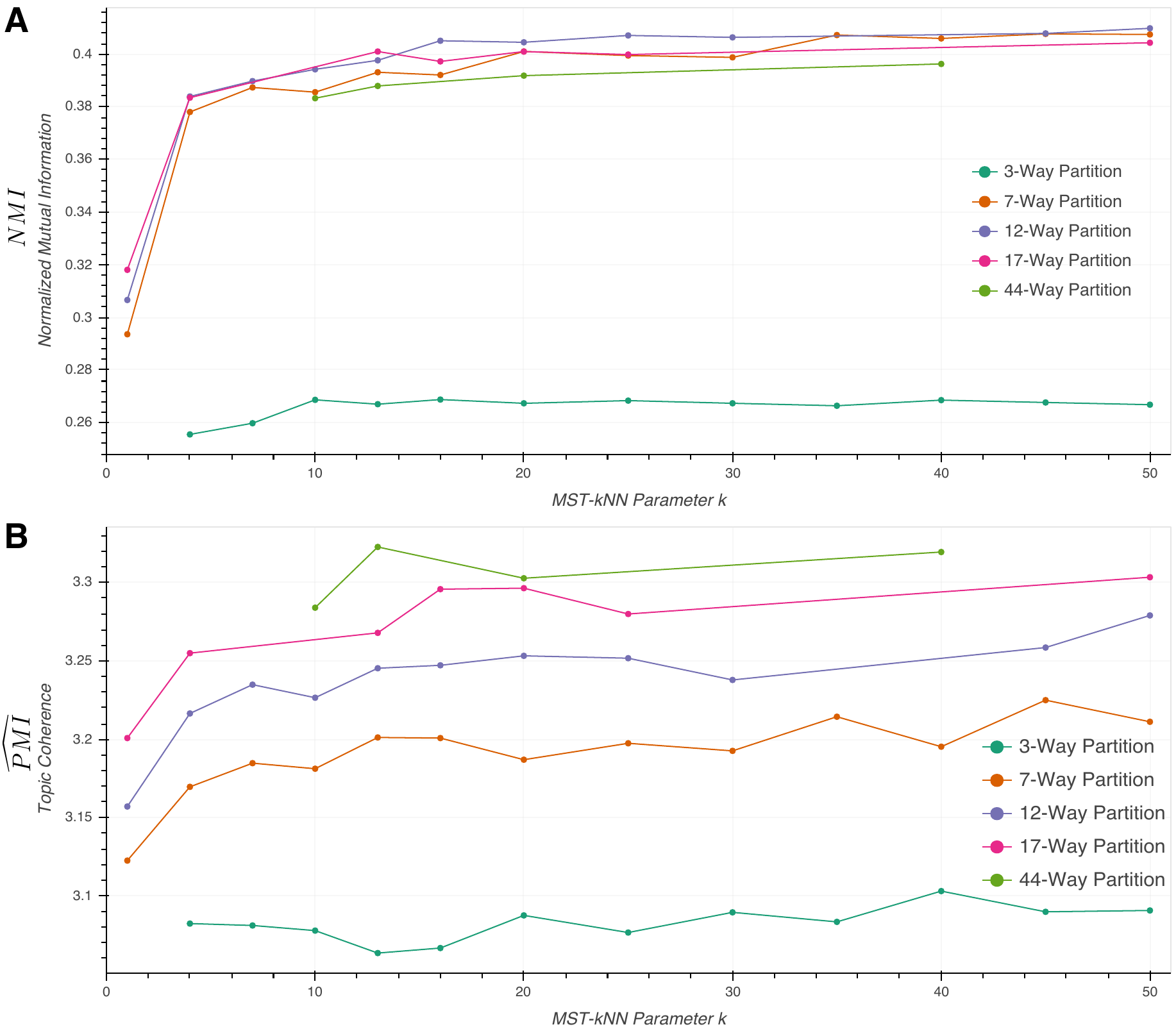}
\caption[Effect of the sparsification]{Effect of the sparsification of the MST-kNN graphs on MS clusters. 
(A) Similarity against the externally hand-coded categories measured with $NMI$; (B) Intrinsic topic coherence of the computed clusters measured with $\widehat{PMI}$.
The clusters have similar quality for values of $k$ above 13-16.}
\label{fig:k_values_SI}
\end{figure}

We sparsify the matrix of cosine similarities using the MST-kNN graph construction. The smaller the value of $k$, the sparser the graph. Sparser graphs have computational advantages for community detection algorithms, but too much sparsification degrades the results.
Figure~\ref{fig:k_values_SI} shows the effect of sparsification in the graph construction on the performance of MS clusters.
Our results are robust to the choice of $k$, provided it is not too small: both the $NMI$ and $\widehat{PMI}$ scores reach a similar level for values of $k$ above 13-16.  Due to computational efficiency, we favour a relatively small value of $k=13$.

\runinhead{Comparison of MS partitions to Latent Dirichlet Allocation with Bag-of-Words (LDA-BoW):}
We have compared the MS results to LDA, a widely used methodology for text analysis. A key difference in LDA is that a different model needs to be trained when the number of topics changes, whereas our MS method produces clusterings at all levels of resolution in one go. 
To compare the outcomes, we trained five LDA models corresponding to the five MS levels in Figure~\ref{fig:MS}. Table~\ref{table:MSvsLDA_all} shows that MS and LDA give partitions that are comparably similar to the hand-coded categories (as measured with $NMI$), with some differences depending on the scale, whereas the MS clusters have higher topic coherence (as given by $\widehat{PMI}$) across all scales.

\begin{table}[htb]
\begin{center}
\begin{tabular}{c||c|c||c|c|}
 & \multicolumn{2}{c||}{\begin{tabular}[c]{@{}c@{}}Similarity to hand-coded \\ categories  ($NMI$)\end{tabular}}
 & \multicolumn{2}{c|}{\begin{tabular}[c]{@{}c@{}} Topic Coherence \\ ($\widehat{PMI}$)\end{tabular}} \\ \hline
 \multicolumn{1}{c||}{\textbf{\begin{tabular}[c]{@{}c@{}}No. of\\ topics/clusters\end{tabular}}} & 
\multicolumn{1}{c|}{LDA} & \multicolumn{1}{c||}{MS} & 
\multicolumn{1}{c|}{LDA} & \multicolumn{1}{c|}{MS} \\ 
\hline
3 	& \textbf{0.311} 	& 0.267 			& 2.991	 	& \textbf{3.033}		\\ \hline
7 	& \textbf{0.409} 	& 0.393 			& 3.218	 	& \textbf{3.303}		\\ \hline
12	& 0.361 			& \textbf{0.398} 	& 3.270		& \textbf{3.517}		\\ \hline
17	& 0.390 			& \textbf{0.401} 	& 3.419		& \textbf{3.457}		\\ \hline
44	& \textbf{0.395} 	& 0.388 			& 3.549		& \textbf{3.716}		\\ \hline
\end{tabular}
\caption{Similarity to hand-coded categories ($NMI$) and topic coherence ($\widehat{PMI}$) for the five MS resolutions in Figure~\ref{fig:MS} and their corresponding LDA models.}
\label{table:MSvsLDA_all}
\end{center}
\end{table}

To give an indication of computational cost, we ran both methods on the same servers. Our method takes approximately 13 hours in total (11 hours to train the Doc2Vec model on 13 million records and 2 hours to produce the full MS scan with 400 partitions across all resolutions). The time required to train just the 5 LDA models on the same corpus amounts to 30 hours (with timings ranging from $\sim$2 hours for the 3 topic LDA model to 12.5 hours for the 44 topic LDA model).  This comparison also highlights the conceptual difference between our multi-scale methodology and LDA topic modelling. While LDA computes topics at a pre-determined level of resolution, our method obtains partitions at all resolutions in one sweep of the Markov time, from which relevant partitions are chosen based on their robustness. The MS partitions at all resolutions are available for further investigation if so needed.

\begin{figure}[htb]
\centering
\includegraphics[width=0.9\textwidth]{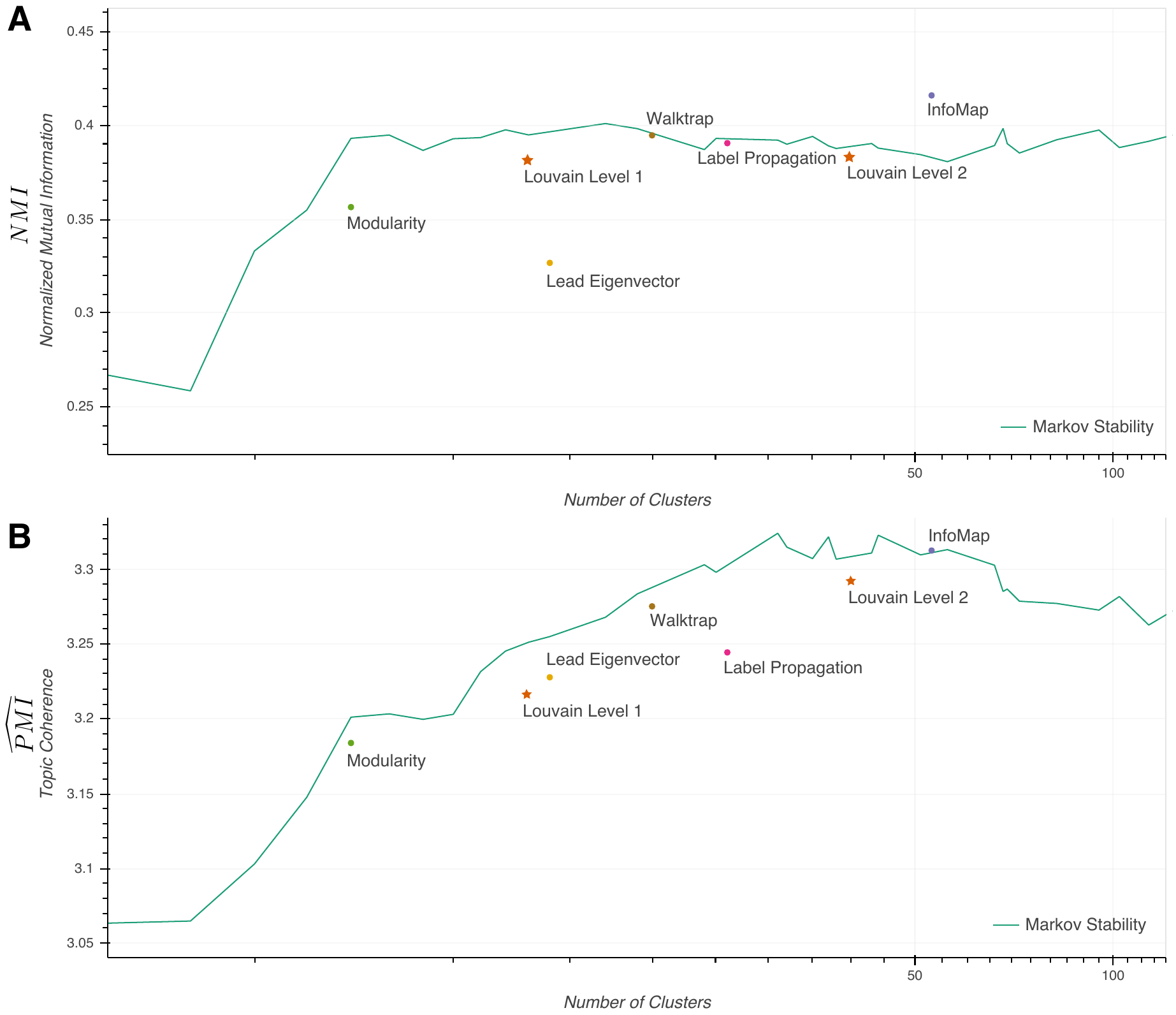}
\caption{Comparison of MS results versus other common community detection or graph partitioning methods: (A) Similarity against the externally hand-coded categories measured with $NMI$; (B) intrinsic topic coherence of the computed clusters measured with $\widehat{PMI}$. MS provides high quality clusters across all scales. } 
\label{fig:commDetvsMS_SI}
\end{figure}

\runinhead{Comparison of MS to other partitioning and community detection algorithms:}
We have partitioned the same kNN-MST graph using several well-known algorithms readily available in code libraries (i.e., the iGraph module for Python): Modularity Optimisation~\citep{modularity_igraph}, InfoMap~\citep{infomap_EPJS},  Walktrap~\citep{walktrap}, Label Propagation~\citep{labelprop}, and Multi-resolution Louvain~\citep{louvain}.  
Note that, in contrast with our multiscale MS analysis, these methods
give just one partition at a particular resolution (or two for the Louvain implementation in iGraph). 
Figure~\ref{fig:commDetvsMS_SI} shows that MS provides improved or equal results to all those other graph partitioning methods for both $NMI$ and $\widehat{PMI}$ across all scales. Only for very fine resolution (more than 50 clusters) does Infomap, which partitions graphs into small clique-like 
subgraphs~\citep{Schaub2012ZoomingLens,schaub2012encoding}, provide a slightly improved $NMI$. Therefore, MS finds both relevant and high quality clusterings across all scales by sweeping the Markov time parameter.

    \section{Using free-text descriptions to predict the degree of harm of patient safety incidents with a supervised classifier} 
    \label{sec:ApplicationEloise}

Here we approach the task of training a supervised classifier that predicts the degree of harm of an incident based on other features of the record (such as location, external category, and medical specialty) and on the textual component of the report. To this end, we use the embedded text vectors and MS cluster labels of the records as features to predict the degree of harm to the patient. 

Each NRLS record has more than 170 features filled manually by healthcare staff, including the degree of harm (DoH) to the patient, a crucial assessment of the reported incident. The incident is classified into five levels: 'No harm', 'Low', 'Moderate', 'Severe', and 'Death'.
However, the reported DoH is not consistent across hospitals and can be unreliable~\citep{riskerror}. 

The lack of reliability of the recorded DoH poses a challenge when training supervised models. Given the size of the dataset, it is not realistic to ask medics to re-evaluate incidents manually.  Instead, we use the publicly available `Learning from mistakes league table' based on NHS staff survey data to identify organisations (NHS Trusts) with `outstanding' (O) and `poor reporting culture' (PRC). 
Our hypothesis is that training our classifiers on records from organisations with better rankings in the league table should lead to improved prediction.
If there is a real disparity in the manual classification among organisations, only incidents labelled by O-ranked Trusts should be regarded as a `ground truth'. 

\subsection{Supervised classification of degree of harm}
We study NRLS incidents reported between 2015 and 2017 from O-ranked and PRC-ranked Trusts. 
The 2015-17 NRLS dataset is very unbalanced: there are 2,038,889 ``No harm'' incidents against only 6,754 ``Death'' incidents. 
To tackle this issue, we sample our dataset as recommended by \citep{ref1}, 
and randomly select 1,016 records each of `No harm' , `Low', and `Moderate', and 508 records each of `Severe' and `Death' incidents, from each type of Trust. We thus obtain two datasets (O and PRC) consisting of a total of 4,064 incidents each.

For each dataset (O and PRC), we train three classifiers (Ridge, Support Vector Machine with a linear kernel, and Random Forest) with five-fold cross validation, and we compute the F-1 scores of each fold to evaluate the model performance.
We first train models using three categories from the reports: location (L), external hand-coded category (C), and medical specialty (S).
We also compute the performance of models trained on text features, both TF-iDF and Doc2Vec. We also study models trained on a mixture of text and categories.
Finally, we run Markov Stability as described above to obtain cluster labels for each dataset (O and PRC) at different resolutions (70, 45, 30 and 13 communities). We then evaluate if it is advantageous to include the labels of the MS clusters as additional features.

\begin{table*}[t!]
\resizebox{\textwidth}{!}
{
\begin{tabular}{l c cccccccccccc}
& & \multicolumn{12}{c} {\textbf{Combinations of features}} \\
\cline{3-14}
& & 1 & 2 & 3 & 4 & 5 & 6 & 7 & 8 &9&10&11&12 \\

\textbf{Dataset}&\textbf{Classifier} & \textbf{L,C,S} & \textbf{TF-iDF}   & \textbf{TF-iDF}   & \textbf{TF-iDF}   & \textbf{TF-iDF} & \textbf{TF-iDF}   & \textbf{TF-iDF}  & \textbf{TF-iDF}  & \textbf{TF-iDF}  & \textbf{TF-iDF}  & \textbf{Doc2Vec} & \textbf{Doc2Vec} \\
&  & & & \multicolumn{1}{c}{\textbf{L}} & \multicolumn{1}{c}{\textbf{C}} & \multicolumn{1}{c}{\textbf{S}} & \multicolumn{1}{c}{\textbf{L,C,S}}  & 
\multicolumn{1}{c}{\textbf{MS-70}} & \multicolumn{1}{c}{\textbf{MS-45}} & \multicolumn{1}{c}{\textbf{MS-30}} & \multicolumn{1}{c}{\textbf{MS-13}} &
\multicolumn{1}{c}{\textbf{}} & \multicolumn{1}{c}{\textbf{MS-30}} \\
\hline
&Ridge & 0.344 & 0.542 & 0.561 & 0.560 & 0.562 & 0.560  & 0.586 & 0.585 & 0.578 & 0.576 & 0.545 & 0.648  \\
O & SVM-LK  & 0.316 & 0.552  & 0.560 & 0.566 & 0.561 & 0.567  & 0.593 & 0.584 & 0.582 & 0.577 & 0.563 & \textbf{0.657}  \\
 & Random Forest & 0.451 & 0.529  & 0.537 & 0.540 & 0.534 & 0.543  & 0.584 & 0.584 & 0.582 & 0.530 & 0.439 & 0.594  \\
\hline 
\\

&Ridge & 0.232 & 0.478 & 0.490 & 0.494 & 0.492 & 0.489 & 0.530 & 0.533 & 0.533 & 0.521 & 0.528 & 0.631  \\
PRC & SVM-LK  & 0.204 & 0.484  & 0.485 & 0.486 & 0.483 & 0.485  & 0.537 & 0.539 & 0.541 & 0.529 & 0.540 & \textbf{0.643}  \\
& Random Forest & 0.379 & 0.474  & 0.473 & 0.479 & 0.473 & 0.479 & 0.497& 0.504& 0.507& 0.500 & 0.419 & 0.573 \\
\hline 
\end{tabular}
}
\caption { Weighted F1-scores for three classifiers (Ridge, SVM with a linear kernel, Random Forest) trained on the O and PRC datasets of incident reports, for different 
features: non-textual categorical features  (L: Localisation; C: Hand-coded Category; S: Medical Specialty); TF-iDF textual features (TF-iDF embedding of text in incident report); Doc2Vec textual features (Doc2Vec embedding of text in incident report); labels of X=70, 45, 30, 13 communities obtained from unsupervised Markov Stability analysis (MS-x). The SVM classifier performs best across the dataset. The classification is better for O-ranked records compared to PRC-ranked records. Text classifiers have highly improved performance compared to purely categorical classifiers. The best classifier is based on Doc2Vec features augmented by MS labels obtained with our unsupervised framework.}
\label{res-exp}
\end{table*}

Table~\ref{res-exp} presents the results of our numerical experiments.
Our first observation is that, for this data, SVM with linear kernel has the best performance (similar to Ridge), and Random Forests perform poorly in general.  
There are several conclusions from our study. First, there is a consistent difference between the scores of the O and PRC datasets (ranging from 1.7\% to 11.2\% for an average of 5.6\%), thus confirming our hypothesis that automated classification performs better when training with data from organizations with better rankings in the league table. 
Second, using text features is highly advantageous in predicting the degree of harm compared to category alone: there is a substantial increase of up to 100\% in the F1 score between column 1 (all three categories) and column 2 (Tf-iDF). Furthermore, adding categorical features (L, C, or S) to the TF-iDF text features improves the scores only marginally (around 2\%), as seen by comparing columns 3--6 with column 2.

Given the demonstrated importance of text, we studied the effect of using more refined textual features for classification.
In columns 7-10, we considered the effect of adding to TF-iDF the MS labels extracted from our text analysis (as described above), and we find a larger improvement of around 7\% with respect to mere TF-iDF (column 2). The improvement is larger for finer clusterings into 70 and 45 communities, which contain enough detail that can be associated with levels of risk (e.g., type of accident). This supports the value of the multi-resolution groupings we have extracted through our analysis. 

We also studied the impact of using Doc2Vec vectors as features. Interestingly, the comparison between columns 2 and 11 shows that there is only a slight improvement of 2\% when using Doc2Vec instead of TF-iDF features for the case of records from O-ranked institutions, but the improvement is of 12\% for the records from PRC Trusts. This differences suggests that the usage of terms is more precise in O-ranked hospitals so that the differences between TF-iDF are minimised, while the advantages of the syntactic and semantic reconstruction of the Doc2Vec embedding becomes more important in the case of PRC Trusts.

\begin{figure}[htb]
\centering
\includegraphics[width=0.8\textwidth]{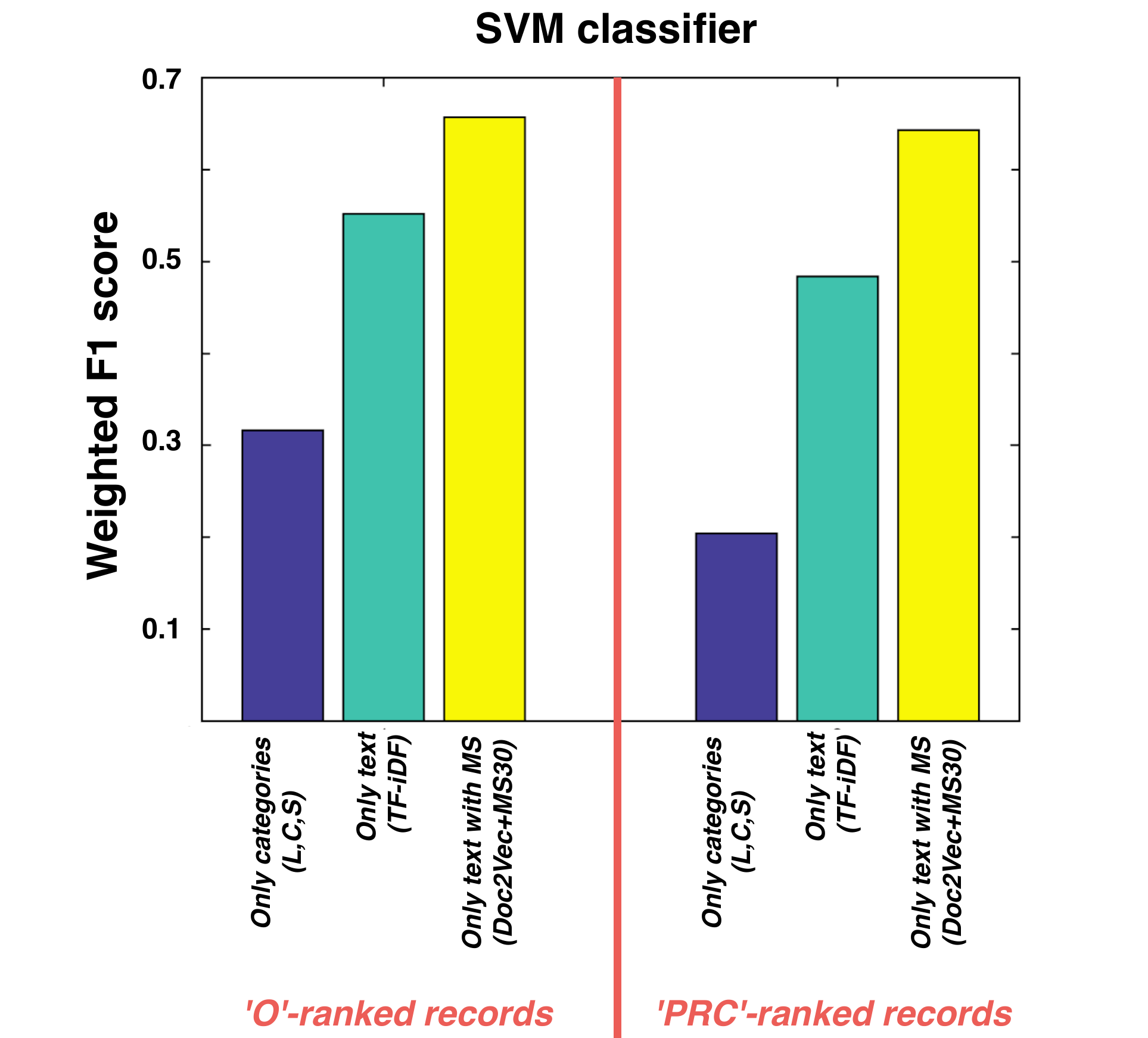}
\caption{Performance of the SVM classifier based on categorical features alone, text features (TF-iDF) alone, and text features (Doc2Vec) with content labels (MS30) on both sets of incident reports: the one collecged from Outstanding Trusts ('O'-ranked) and from Trusts with a Poor Reporting Culture ('PRC'-ranked). The inclusion of more sophisticated text and content labels improves prediction and closes the gap in the quality between both sets of records.} 
\label{fig:SVM_comparison}
\end{figure}

Based on these findings, we build our final model that uses a Support Vector Machine classifier with both Doc2Vec embeddings and the MS labels for 30 content clusters (encoded via a One-Hot encoder) as features. We choose to keep only 30 communities as this performs well when combined with the Doc2Vec embedding (without slowing too much the classifier). We performed a grid search to optimise the hyperparameters of our model (penalty = 10, tolerance for stopping criterion = 0.0001, linear kernel). For the O-ranked records, our model achieves a weighted F1 score of 0.657, with a
19\% improvement with respect to TF-iDF text features and a 107\% improvement with respect to categorical features.
(For the PRC records, the corresponding improvements are 33\% and 215\%, respectively.) Note that similar improvements are also obtained for the other classifiers when using Doc2Vec and MS labels as features. It is also worth noting that the differences in the prediction of DoH between PRC and O-ranked records is reduced when using text tools and, specifically, the F1-score of the SVM classifier based on Doc2Vec with MS is almost the same for both datasets. Hence the difference in the quality of the reporting categories can be ameliorated by the use of the textual content of the reports.
We summarise the main comparison of the performance of the SVM classifier based on categorical, raw text, and text with content for both datasets in Figure~\ref{fig:SVM_comparison}.

\runinhead{Examination of the types of errors and \textit{ex novo} re-classification by clinicians:}

A further analysis of the confusion matrices used to compute the F1 score reveals that most of the errors of our model are concentrated in the `No harm', `Low harm' and `Moderate harm' categories, whereas fewer errors are incurred in the `Severe harm' and `Death' categories. Therefore, our method is more likely to return  false alarms rather than missing important and harmful incidents.

In order to have a further evaluation of our results, we asked three clinicians to analyse \textit{ex novo} a randomly chosen sample of 135 descriptions of incidents, and to determine their degree of harm based on the information in the incident report. The sample was selected from the O-ranked dataset and no extra information apart from the text was provided. We then compared the DoH assigned by the clinicians with both the results of our classifier and the recorded DoH in the dataset. 

Remarkably, the agreement rate of the clinicians' assessment with the recorded DoH was surprisingly low. For example, the agreement in the `No Harm' incidents was only 38\%,  and in  the `Severe' incidents only 49\%. In most cases, though, the disparities amounted to switching the DoH by one degree above or below. 
To reduce this variability, we analysed the outcomes in terms of three larger groups: `No Harm' and `Low Harm' incidents were considered as one outcome; `Moderate Harm' was kept separate; and `Severe Harm' and `Death' were grouped as one outcome, since they both need to be notified to NHS safety managers.

The results are presented in Table~\ref{res-med}.
Our classification agrees as well as the pre-existing DoH in the dataset with the \textit{ex novo} assessment of the clinicians, but our method has higher agreement in the severe and deadly incidents. 
These results confirm 
that our method performs as well as the original annotators but is 
better at identifying risky events.

\begin{table}[h!]
\centering
\textbf{Agreement with clinicians' \textit{ex novo} classification} \\
\begin{tabular}{|c|c|c|}
\cline{1-3}
Type of  & Pre-existing classification & Prediction of  \\ 
 incident & in dataset & our model \\  \hline
\multicolumn{1}{|c|}{`No Harm' and `Low Harm'}  & 59 \% & 54 \%  \\ \hline
\multicolumn{1}{|c|}{`Moderate Harm'}& 49 \% & 41 \%  \\ \hline
\multicolumn{1}{|c|}{`Severe Harm' and `Death'} & 66 \% & 72 \%  \\ \hline \hline
\multicolumn{1}{|c|}{\textit{Average}} & \textit{59 \%} & \textit{58 \%} \\ \hline 
\end{tabular}
\caption {The \textit{ex novo} re-classification by three clinicians of 135 incident reports (chosen at random) is compared to the pre-existing classification in the dataset and the prediction of our model.}
\label{res-med} 
\end{table}

    \section{Discussion}
    \label{sec:Discussion}

We have applied a multiscale graph partitioning algorithm (Markov Stability) to extract content-based clusters of documents from a textual dataset of incident reports in an unsupervised manner at different levels of resolution. The method uses paragraph vectors to represent the records and analyses the ensuing  similarity graph of documents through multi-resolution capabilities to capture clusters without imposing \textit{a priori} their number or structure. The different levels of resolution found to be relevant can be chosen by the practitioner to suit the requirements of detail for each specific task. 
For example, the top level categories of the pre-defined classification hierarchy are highly diverse in size, with large groups such as `Patient accident', `Medication', `Clinical assessment', `Documentation', `Admissions/Transfer' or `Infrastructure' alongside small, specific groups such as `Aggressive behaviour', `Patient abuse', `Self-harm' or `Infection control'. Our multi-scale partitioning finds additional subcategories with medical detail within some of the large categories (Fig.~\ref{fig:44comms}~and~\ref{fig:44comms_SI}).

Our \textit{a posteriori} analysis showed that the method recovers meaningful clusters of content as measured by the similarity of the groups against the hand-coded categories and by the intrinsic topic coherence of the clusters.
The clusters have high medical content, thus providing complementary information to the externally imposed classification categories. Indeed, some of the most relevant and persistent communities emerge because of their highly homogeneous medical content, even if they cannot be mapped to standardised external categories. 

An area of future research will be to confirm if the finer unsupervised cluster found by our analysis are consistent with a second level in the hierarchy of external categories (Level 2, around 100 categories), which is used less consistently in hospital settings. 
The use of content-driven classification of reports could also be important within current efforts by the World Health Organisation (WHO) under the framework for the International Classification for Patient Safety (ICPS)~\citep{who_ICPS}
to establish a set of conceptual categories to monitor, analyse and interpret information to improve patient care. 

We have used our clusters within a supervised classifier to predict the degree of harm of an incident based only on free-text descriptions. 
The degree of harm is an important measure in hospital evaluation and has been shown to depend on the reporting culture of the particular organisation.
Overall, our method shows that text description complemented by the topic labels extracted by our method show improved performance in this task.
The use of such enhanced NLP tools could help improve reporting frequency and quality, in addition to reducing burden to staff, since most of the necessary information can be retrieved automatically from text descriptions. Further work, would aim to add interpretability to the supervised classification~\citep{lime}, so as to provide medical staff with a clearer view of the outcomes of our method and to encourage its uptake. 

One of the advantages of a free text analytical approach is the provision, in a timely manner, of an intelligible description of incident report categories derived directly from the 'words' of the reporters themselves. 
Insights from the analysis of such free text entries can add rich information than would have not otherwise been obtained from pre-defined classes. 
Not only could this improve the current state of play where much of the free text of these reports goes unused, but by avoiding the strict assignment to pre-defined categories of fixed granularity free text analysis could open an opportunity for feedback and learning through more nuanced classifications as a complementary axis to existing approaches.

Currently, local incident reporting systems used by hospitals to submit reports to the NRLS require risk managers to improve data quality, due to errors or uncertainty in categorisation. The application of free text analytical approaches has the potential to free up time from this labour-intensive task, focussing instead in quality improvement derived from the content of the data itself. Additionally, the method allows for the discovery of emerging topics or classes of incidents directly from the data when such events do not fit existing categories by using methods for anomaly detection to decide whether new topic clusters should be created. This is a direction of future work. 

Further work also includes the use of our method to enable comparisons across healthcare organisations and also to monitor changes in their incident reports over time. Another interesting direction is to provide online classification suggestions to users based on the text they input as an aid with decision support and data collection, which can also help fine-tune the predefined categories.
Finally, it would be interesting to test if the use of deep learning algorithms can improve our classification scores.

\begin{acknowledgement}
We thank 
Elias Bamis, Zijing Liu and Michael Schaub for helpful discussions.
This research was supported by the National Institute for Health Research (NIHR) Imperial Patient Safety Translational Research Centre and NIHR Imperial Biomedical Research Centre. The views expressed are those of the authors and not necessarily those of the NHS, the NIHR, or the Department of Health. 
All authors acknowledge support from the EPSRC through award EP/N014529/1 funding the EPSRC Centre for Mathematics of Precision Healthcare. 
\end{acknowledgement}

\bibliographystyle{spbasic}
\bibliography{references}

\begin{thebibliography}{58}
\providecommand{\natexlab}[1]{#1}
\providecommand{\url}[1]{{#1}}
\providecommand{\urlprefix}{URL }
\expandafter\ifx\csname urlstyle\endcsname\relax
  \providecommand{\doi}[1]{DOI~\discretionary{}{}{}#1}\else
  \providecommand{\doi}{DOI~\discretionary{}{}{}\begingroup
  \urlstyle{rm}\Url}\fi
\providecommand{\eprint}[2][]{\url{#2}}

\bibitem[{Aggarwal and Zhai(2012)}]{book}
Aggarwal CC, Zhai C (2012) Mining text data. Springer Science \& Business Media

\bibitem[{Agirre et~al.(2016)Agirre, Banea, Cer, Diab, Gonzalez-Agirre,
  Mihalcea, Rigau, and Wiebe}]{semevalSTS2016}
Agirre E, Banea C, Cer D, Diab M, Gonzalez-Agirre A, Mihalcea R, Rigau G, Wiebe
  J (2016) {Semeval-2016 task 1: Semantic textual similarity, monolingual and
  cross-lingual evaluation}. In: Proceedings of the 10th International Workshop
  on Semantic Evaluation (SemEval-2016), pp 497--511

\bibitem[{Altuncu et~al.(2018)Altuncu, Yaliraki, and
  Barahona}]{altuncu_content-driven_2018}
Altuncu MT, Yaliraki SN, Barahona M (2018) Content-driven, unsupervised
  clustering of news articles through multiscale graph partitioning.
  arXiv:180801175 [cs, math] \urlprefix\url{http://arxiv.org/abs/1808.01175},
  arXiv: 1808.01175

\bibitem[{Altuncu et~al.(2019)Altuncu, Mayer, Yaliraki, and Barahona}]{tarik}
Altuncu MT, Mayer E, Yaliraki SN, Barahona M (2019) From free text to clusters
  of content in health records: an unsupervised graph partitioning approach.
  Applied Network Science 4(1):2

\bibitem[{Bacik et~al.(2016)Bacik, Schaub, Beguerisse-D{\'{i}}az, Billeh, and
  Barahona}]{bacik_celegans}
Bacik KA, Schaub MT, Beguerisse-D{\'{i}}az M, Billeh YN, Barahona M (2016)
  {Flow-Based Network Analysis of the Caenorhabditis elegans Connectome}. PLOS
  Computational Biology 12(8):1--27, \doi{10.1371/journal.pcbi.1005055},
  \urlprefix\url{https://doi.org/10.1371/journal.pcbi.1005055}

\bibitem[{Beguerisse-Diaz et~al.(2013)Beguerisse-Diaz, Vangelov, and
  Barahona}]{rbs_rmst}
Beguerisse-Diaz M, Vangelov B, Barahona M (2013) {Finding role communities in
  directed networks using Role-Based Similarity, Markov Stability and the
  Relaxed Minimum Spanning Tree}. In: 2013 IEEE Global Conference on Signal and
  Information Processing, GlobalSIP 2013 - Proceedings, pp 937--940,
  \doi{10.1109/GlobalSIP.2013.6737046}

\bibitem[{Beguerisse-D{\'{i}}az et~al.(2014)Beguerisse-D{\'{i}}az,
  Gardu{\~{n}}o-Hern{\'{a}}ndez, Vangelov, Yaliraki, and Barahona}]{ukRiotsTw}
Beguerisse-D{\'{i}}az M, Gardu{\~{n}}o-Hern{\'{a}}ndez G, Vangelov B, Yaliraki
  SN, Barahona M (2014) {Interest communities and flow roles in directed
  networks: the Twitter network of the UK riots.} Journal of the Royal Society,
  Interface / the Royal Society 11(101):20140940, \doi{10.1098/rsif.2014.0940}

\bibitem[{Bird et~al.(2009)Bird, Klein, and Loper}]{nltk}
Bird S, Klein E, Loper E (2009) {Natural Language Processing with Python}, 1st
  edn. O'Reilly Media, Inc.

\bibitem[{Blei et~al.(2003)Blei, Ng, and Jordan}]{lda}
Blei DM, Ng AY, Jordan MI (2003) {Latent Dirichlet Allocation}. Journal of
  Machine Learning Research pp 993--1022, \doi{10.1162/jmlr.2003.3.4-5.993},
  \urlprefix\url{http://www.jmlr.org/papers/volume3/blei03a/blei03a.pdf}

\bibitem[{Blondel et~al.(2008)Blondel, Guillaume, Lambiotte, and
  Lefebvre}]{louvain}
Blondel VD, Guillaume JL, Lambiotte R, Lefebvre E (2008) {Fast unfolding of
  communities in large networks}. Journal of Statistical Mechanics: Theory and
  Experiment 2008(10):P10008, \doi{10.1088/1742-5468/2008/10/P10008}

\bibitem[{Cer et~al.(2017)Cer, Diab, Agirre, Lopez-Gazpio, and
  Specia}]{semevalSTS2017}
Cer D, Diab M, Agirre E, Lopez-Gazpio I, Specia L (2017) {SemEval-2017 Task 1:
  Semantic Textual Similarity-Multilingual and Cross-lingual Focused
  Evaluation}. arXiv preprint arXiv:170800055

\bibitem[{Clauset et~al.(2004)Clauset, Newman, and Moore}]{modularity_igraph}
Clauset A, Newman MEJ, Moore C (2004) {Finding community structure in very
  large networks}. Physical review E 70(6):66111

\bibitem[{Colijn et~al.(2017)Colijn, Jones, Johnston, Yaliraki, and
  Barahona}]{colijn2017toward}
Colijn C, Jones N, Johnston IG, Yaliraki S, Barahona M (2017) {Toward precision
  healthcare: Context and mathematical challenges}. Frontiers in Physiology
  8(MAR):136, \doi{10.3389/fphys.2017.00136},
  \urlprefix\url{https://spiral.imperial.ac.uk:8443/bitstream/10044/1/44892/11/fphys-08-00136.pdf}

\bibitem[{Dai et~al.(2015)Dai, Olah, and Le}]{dai2015document}
Dai AM, Olah C, Le QV (2015) {Document Embedding with Paragraph Vectors}. CoRR
  abs/1507.0, \urlprefix\url{http://arxiv.org/abs/1507.07998}

\bibitem[{Delvenne et~al.(2010)Delvenne, Yaliraki, and
  Barahona}]{pnasStability}
Delvenne JC, Yaliraki SN, Barahona M (2010) {Stability of graph communities
  across time scales.} Proceedings of the National Academy of Sciences of the
  United States of America 107(29):12755--60, \doi{10.1073/pnas.0903215107},
  \urlprefix\url{http://www.ncbi.nlm.nih.gov/pubmed/20615936
  http://www.pubmedcentral.nih.gov/articlerender.fcgi?artid=PMC2919907}

\bibitem[{Delvenne et~al.(2013)Delvenne, Schaub, Yaliraki, and
  Barahona}]{Delvenne2013}
Delvenne JC, Schaub MT, Yaliraki SN, Barahona M (2013) {The Stability of a
  Graph Partition: A Dynamics-Based Framework for Community Detection}. In:
  Mukherjee A, Choudhury M, Peruani F, Ganguly N, Mitra B (eds) Dynamics On and
  Of Complex Networks, Volume 2: Applications to Time-Varying Dynamical
  Systems, Springer New York, New York, NY, pp 221--242,
  \doi{10.1007/978-1-4614-6729-8{\_}11},
  \urlprefix\url{https://doi.org/10.1007/978-1-4614-6729-8_11}

\bibitem[{Fang et~al.(2016)Fang, Macdonald, Ounis, and
  Habel}]{twitter_pmi_coherence}
Fang A, Macdonald C, Ounis I, Habel P (2016) {Topics in Tweets: A User Study of
  Topic Coherence Metrics for Twitter Data}. In: Ferro N, Crestani F, Moens MF,
  Mothe J, Silvestri F, Di~Nunzio GM, Hauff C, Silvello G (eds) Advances in
  Information Retrieval, Springer International Publishing, Cham, pp 492--504

\bibitem[{Friedman et~al.(2008)Friedman, Hastie, and Tibshirani}]{g_lasso}
Friedman J, Hastie T, Tibshirani R (2008) {Sparse inverse covariance estimation
  with the graphical lasso}. Biostatistics 9(3):432--441

\bibitem[{Hashimoto et~al.(2016)Hashimoto, Kontonatsios, Miwa, and
  Ananiadou}]{Hashimoto2016TopicReviews}
Hashimoto K, Kontonatsios G, Miwa M, Ananiadou S (2016) {Topic detection using
  paragraph vectors to support active learning in systematic reviews}. Journal
  of Biomedical Informatics 62:59--65, \doi{10.1016/J.JBI.2016.06.001},
  \urlprefix\url{https://www.sciencedirect.com/science/article/pii/S1532046416300442}

\bibitem[{Jacomy et~al.(2014)Jacomy, Venturini, Heymann, and
  Bastian}]{forceAtlas2}
Jacomy M, Venturini T, Heymann S, Bastian M (2014) {ForceAtlas2, a continuous
  graph layout algorithm for handy network visualization designed for the Gephi
  software}. PLoS ONE 9(6), \doi{10.1371/journal.pone.0098679}

\bibitem[{Jones et~al.(2001)Jones, Oliphant, Peterson, and {others}}]{scipy}
Jones E, Oliphant T, Peterson P, {others} (2001) {{\{}SciPy{\}}: Open source
  scientific tools for {\{}Python{\}}}. \urlprefix\url{http://www.scipy.org/}

\bibitem[{Lambiotte et~al.(2008)Lambiotte, Delvenne, and
  Barahona}]{lambiotte_arxiv}
Lambiotte R, Delvenne JC, Barahona M (2008) {Laplacian Dynamics and Multiscale
  Modular Structure in Networks}. arXiv preprint arXiv:08121770 pp 1--29,
  \doi{10.1109/TNSE.2015.2391998},
  \urlprefix\url{http://arxiv.org/abs/0812.1770%5Cnhttp://dx.doi.org/10.1109/TNSE.2015.2391998}

\bibitem[{Lambiotte et~al.(2014)Lambiotte, Delvenne, and
  Barahona}]{LambiotteMarkovProcess}
Lambiotte R, Delvenne JC, Barahona M (2014) {Random Walks, Markov Processes and
  the Multiscale Modular Organization of Complex Networks}. IEEE Transactions
  on Network Science and Engineering 1(2):76--90,
  \doi{10.1109/TNSE.2015.2391998}

\bibitem[{Lancichinetti et~al.(2015)Lancichinetti, Sirer, Wang, Acuna,
  K{\"{o}}rding, and Amaral}]{PhysRevX.5.011007}
Lancichinetti A, Sirer MI, Wang JX, Acuna D, K{\"{o}}rding K, Amaral LAN (2015)
  {High-Reproducibility and High-Accuracy Method for Automated Topic
  Classification}. Phys Rev X 5(1):11007, \doi{10.1103/PhysRevX.5.011007},
  \urlprefix\url{https://link.aps.org/doi/10.1103/PhysRevX.5.011007}

\bibitem[{Lau et~al.(2011)Lau, Grieser, Newman, and Baldwin}]{pmi_coherence2}
Lau J, Grieser K, Newman D, Baldwin T (2011) {Automatic labeling of topic
  models}. In: Proceedings of the 49th Annual Meeting of the Association for
  Computational Linguistics: Human Language Technologies - Volume 1,
  Association for Computational Linguistics, Portland, Oregon, pp 1536--1545,
  \doi{10.1109/JSEN.2012.2223209},
  \urlprefix\url{http://www.aclweb.org/anthology/P11-1154
  http://dl.acm.org/citation.cfm?id=2002472.2002658}

\bibitem[{Lau and Baldwin(2016)}]{jhlau}
Lau JH, Baldwin T (2016) {An Empirical Evaluation of doc2vec with Practical
  Insights into Document Embedding Generation}. In: Proceedings of the 1st
  Workshop on Representation Learning for NLP, Rep4NLP@ACL 2016, Berlin,
  Germany, August 11, 2016, pp 78--86, \doi{10.18653/v1/W16-1609},
  \urlprefix\url{https://doi.org/10.18653/v1/W16-1609}

\bibitem[{Le and Mikolov(2014)}]{d2v_mikolov}
Le Q, Mikolov T (2014) {Distributed Representations of Sentences and
  Documents}. In: Proceedings of the 31st International Conference on
  International Conference on Machine Learning - Volume 32, JMLR.org, ICML'14,
  pp II--1188--II--1196,
  \urlprefix\url{http://dl.acm.org/citation.cfm?id=3044805.3045025}

\bibitem[{Meil{\u{a}}(2007)}]{Meila2007}
Meil{\u{a}} M (2007) {Comparing clusterings—an information based distance}.
  Journal of Multivariate Analysis 98(5):873--895,
  \doi{10.1016/J.JMVA.2006.11.013},
  \urlprefix\url{https://www.sciencedirect.com/science/article/pii/S0047259X06002016}

\bibitem[{Mikolov et~al.(2013{\natexlab{a}})Mikolov, Corrado, Chen, and
  Dean}]{mikolov2013efficient}
Mikolov T, Corrado G, Chen K, Dean J (2013{\natexlab{a}}) {Efficient Estimation
  of Word Representations in Vector Space}. Proceedings of the International
  Conference on Learning Representations (ICLR 2013) pp 1--12,
  \doi{10.1162/153244303322533223},
  \urlprefix\url{http://arxiv.org/pdf/1301.3781v3.pdf}

\bibitem[{Mikolov et~al.(2013{\natexlab{b}})Mikolov, Sutskever, Chen, Corrado,
  and Dean}]{w2v2}
Mikolov T, Sutskever I, Chen K, Corrado G, Dean J (2013{\natexlab{b}})
  {Distributed Representations of Words and Phrases and Their
  Compositionality}. In: Proceedings of the 26th International Conference on
  Neural Information Processing Systems - Volume 2, Curran Associates Inc.,
  USA, NIPS'13, pp 3111--3119,
  \urlprefix\url{http://dl.acm.org/citation.cfm?id=2999792.2999959}

\bibitem[{Newman et~al.(2009)Newman, Karimi, and Cavedon}]{pmi_coherence}
Newman D, Karimi S, Cavedon L (2009) {External evaluation of topic models}. In:
  Kay J, Thomas P, Trotman A (eds) in Australasian Doc. Comp. Symp., 2009,
  School of Information Technologies, University of Sydney, pp 11--18

\bibitem[{Newman et~al.(2010)Newman, Lau, Grieser, and
  Baldwin}]{Newman:2010:AET:1857999.1858011}
Newman D, Lau JH, Grieser K, Baldwin T (2010) {Automatic Evaluation of Topic
  Coherence}. In: Human Language Technologies: The 2010 Annual Conference of
  the North American Chapter of the Association for Computational Linguistics,
  Association for Computational Linguistics, Stroudsburg, PA, USA, HLT '10, pp
  100--108, \urlprefix\url{http://dl.acm.org/citation.cfm?id=1857999.1858011}

\bibitem[{Newman et~al.(2011)Newman, Bonilla, and Buntine}]{pmi_coherence_lda}
Newman D, Bonilla EV, Buntine W (2011) {Improving Topic Coherence with
  Regularized Topic Models}. In: Shawe-Taylor J, Zemel RS, Bartlett PL, Pereira
  F, Weinberger KQ (eds) Advances in Neural Information Processing Systems 24,
  Curran Associates, Inc., pp 496--504,
  \urlprefix\url{http://papers.nips.cc/paper/4291-improving-topic-coherence-with-regularized-topic-models.pdf}

\bibitem[{NRLS(2014)}]{degree_of_harm}
NRLS (2014) {Degree of harm FAQ - National Reporting and Learning System -
  NHS}. \url
  {https://improvement.nhs.uk/documents/1673/NRLS_Degree_of_harm_FAQs_-_final_v1.1.pdf}

\bibitem[{Ong et~al.(2010)Ong, Magrabi, and Coiera}]{ref2}
Ong MS, Magrabi F, Coiera E (2010) Automated categorisation of clinical
  incident reports using statistical text classification. Qual Saf Health Care
  19(6):e55--e55

\bibitem[{Ong et~al.(2012)Ong, Magrabi, and Coiera}]{ref1}
Ong MS, Magrabi F, Coiera E (2012) Automated identification of extreme-risk
  events in clinical incident reports. Journal of the American Medical
  Informatics Association 19(e1):e110--e118

\bibitem[{Pons and Latapy(2005)}]{walktrap}
Pons P, Latapy M (2005) {Computing communities in large networks using random
  walks}. In: International symposium on computer and information sciences,
  Springer, pp 284--293

\bibitem[{Porter(1980)}]{porter_old}
Porter MF (1980) {An algorithm for suffix stripping}. Program 14(3):130--137,
  \doi{10.1108/eb046814}, \urlprefix\url{https://doi.org/10.1108/eb046814}

\bibitem[{Porter(2006)}]{snowball}
Porter MF (2006) {http://snowballstem.org/}

\bibitem[{Raghavan et~al.(2007)Raghavan, Albert, and Kumara}]{labelprop}
Raghavan UN, Albert R, Kumara S (2007) {Near linear time algorithm to detect
  community structures in large-scale networks}. Physical review E 76(3):36106

\bibitem[{Rehurek and Sojka(2010)}]{gensim}
Rehurek R, Sojka P (2010) {Software Framework for Topic Modelling with Large
  Corpora}. In: Proceedings of the LREC 2010 Workshop on New Challenges for NLP
  Frameworks, ELRA, Valletta, Malta, pp 45--50

\bibitem[{Ribeiro et~al.(2016)Ribeiro, Singh, and Guestrin}]{lime}
Ribeiro MT, Singh S, Guestrin C (2016) "why should {I} trust you?": Explaining
  the predictions of any classifier. In: Proceedings of the 22nd {ACM} {SIGKDD}
  International Conference on Knowledge Discovery and Data Mining, San
  Francisco, CA, USA, August 13-17, 2016, pp 1135--1144

\bibitem[{Rosenberg and Hirschberg(2007)}]{vmeasure}
Rosenberg A, Hirschberg J (2007) {V-measure: A conditional entropy-based
  external cluster evaluation measure}. In: Proceedings of the 2007 joint
  conference on empirical methods in natural language processing and
  computational natural language learning (EMNLP-CoNLL)

\bibitem[{Rosvall et~al.(2009)Rosvall, Axelsson, and Bergstrom}]{infomap_EPJS}
Rosvall M, Axelsson D, Bergstrom CT (2009) {The map equation}. The European
  Physical Journal Special Topics 178(1):13--23

\bibitem[{Rychalska et~al.(2016)Rychalska, Pakulska, Chodorowska, Walczak, and
  Andruszkiewicz}]{semeval2016Samsung}
Rychalska B, Pakulska K, Chodorowska K, Walczak W, Andruszkiewicz P (2016)
  {Samsung Poland NLP Team at SemEval-2016 Task 1: Necessity for diversity;
  combining recursive autoencoders, WordNet and ensemble methods to measure
  semantic similarity.} In: Proceedings of the 10th International Workshop on
  Semantic Evaluation (SemEval-2016), Association for Computational
  Linguistics, pp 602--608, \doi{10.18653/v1/S16-1091},
  \urlprefix\url{http://www.aclweb.org/anthology/S16-1091}

\bibitem[{Schaub et~al.(2012{\natexlab{a}})Schaub, Delvenne, Yaliraki, and
  Barahona}]{Schaub2012ZoomingLens}
Schaub MT, Delvenne JC, Yaliraki SN, Barahona M (2012{\natexlab{a}}) {Markov
  dynamics as a zooming lens for multiscale community detection: Non
  clique-like communities and the field-of-view limit}. PLoS ONE
  \doi{10.1371/journal.pone.0032210}

\bibitem[{Schaub et~al.(2012{\natexlab{b}})Schaub, Lambiotte, and
  Barahona}]{schaub2012encoding}
Schaub MT, Lambiotte R, Barahona M (2012{\natexlab{b}}) {Encoding dynamics for
  multiscale community detection: Markov time sweeping for the map equation}.
  Physical Review E - Statistical, Nonlinear, and Soft Matter Physics 86(2),
  \doi{10.1103/PhysRevE.86.026112},
  \urlprefix\url{https://arxiv.org/pdf/1109.6642.pdf}

\bibitem[{Schaub et~al.(2017)Schaub, Delvenne, Rosvall, and
  Lambiotte}]{Schaub2017}
Schaub MT, Delvenne JC, Rosvall M, Lambiotte R (2017) {The many facets of
  community detection in complex networks}. Applied Network Science 2(1):4,
  \doi{10.1007/s41109-017-0023-6},
  \urlprefix\url{https://doi.org/10.1007/s41109-017-0023-6}

\bibitem[{Schubert et~al.(2017)Schubert, Spitz, Weiler, and Gertz}]{wordClouds}
Schubert E, Spitz A, Weiler M, Gertz JGM (2017) {Semantic Word Clouds with
  Background Corpus Normalization and t-distributed Stochastic Neighbor
  Embedding}. CoRR abs/1708.0

\bibitem[{Sculley and Wachman(2007)}]{svmeg}
Sculley D, Wachman GM (2007) Relaxed online svms for spam filtering. In:
  Proceedings of the 30th annual international ACM SIGIR conference on Research
  and development in information retrieval, ACM, pp 415--422

\bibitem[{Spielman and Srivastava(2011)}]{spielman2011graph}
Spielman DA, Srivastava N (2011) {Graph sparsification by effective
  resistances}. SIAM Journal on Computing 40(6):1913--1926

\bibitem[{Strehl and Ghosh(2003)}]{nmi}
Strehl A, Ghosh J (2003) {Cluster Ensembles-A Knowledge Reuse Framework for
  Combining Multiple Partitions}. J Mach Learn Res 3:583--617,
  \doi{10.1162/153244303321897735},
  \urlprefix\url{https://doi.org/10.1162/153244303321897735}

\bibitem[{Tian et~al.(2017)Tian, Zhou, Lan, and Wu}]{semaval2017ECNU}
Tian J, Zhou Z, Lan M, Wu Y (2017) {ECNU at SemEval-2017 Task 1: Leverage
  Kernel-based Traditional NLP features and Neural Networks to Build a
  Universal Model for Multilingual and Cross-lingual Semantic Textual
  Similarity}. In: Proceedings of the 11th International Workshop on Semantic
  Evaluation (SemEval-2017), Association for Computational Linguistics, pp
  191--197, \doi{10.18653/v1/S17-2028},
  \urlprefix\url{http://www.aclweb.org/anthology/S17-2028}

\bibitem[{Tumminello et~al.(2005)Tumminello, Aste, Di~Matteo, and
  Mantegna}]{pmfg}
Tumminello M, Aste T, Di~Matteo T, Mantegna RN (2005) {A tool for filtering
  information in complex systems.} Proceedings of the National Academy of
  Sciences of the United States of America 102(30):10421--6,
  \doi{10.1073/pnas.0500298102},
  \urlprefix\url{http://www.ncbi.nlm.nih.gov/pubmed/16027373
  http://www.pubmedcentral.nih.gov/articlerender.fcgi?artid=PMC1180754}

\bibitem[{Veenstra et~al.(2017)Veenstra, Cooper, and Phelps}]{mstknn}
Veenstra P, Cooper C, Phelps S (2017) {Spectral clustering using the kNN-MST
  similarity graph}. In: 2016 8th Computer Science and Electronic Engineering
  Conference, CEEC 2016 - Conference Proceedings, Institute of Electrical and
  Electronics Engineers Inc., pp 222--227, \doi{10.1109/CEEC.2016.7835917}

\bibitem[{Willett(2006)}]{porter}
Willett P (2006) {The Porter stemming algorithm: then and now}. Program
  40(3):219--223, \doi{10.1108/00330330610681295},
  \urlprefix\url{https://www.emeraldinsight.com/doi/10.1108/00330330610681295}

\bibitem[{Williams and Ashcroft(2009)}]{riskerror}
Williams SD, Ashcroft DM (2009) Medication errors: how reliable are the
  severity ratings reported to the national reporting and learning system?
  International journal for quality in health care 21(5):316--320

\bibitem[{{World Health Organization {\&} WHO Patient Safety}(2010)}]{who_ICPS}
{World Health Organization {\&} WHO Patient Safety} (2010) {Conceptual
  framework for the international classification for patient safety version
  1.1: final technical report}. Tech. Rep. January, World Health Organization,
  Geneva, \doi{WHO/IER/PSP/2010.2},
  \urlprefix\url{http://www.who.int/iris/handle/10665/70882}

\end{thebibliography}
\end{document}